\documentclass[acmsmall]{acmart}
\usepackage{xspace}
\usepackage{wasysym}

\setcopyright{acmlicensed}
\acmJournal{PACMHCI}
\acmYear{2020} \acmVolume{4} \acmNumber{CSCW2} \acmArticle{131} \acmMonth{10} \acmPrice{15.00}\acmDOI{10.1145/3415202}

\received{January 2020}
\received[revised]{June 2020}
\received[accepted]{July 2020}

\newcommand{\xhdr}[1]{{\noindent\bfseries #1.}}

\newcommand{\expt}[1]{\ensuremath{\mathbb{E}[#1]}\xspace}
\newcommand{\condexpt}[2]{\ensuremath{\expt{#1 \!\mid\! #2}}\xspace}

\newcommand{\inter}[1]{\ensuremath{#1}\xspace}

\newcommand{\vartend}{\ensuremath{\mathcal{T}}\xspace}
\newcommand{\obstend}[1]{\ensuremath{\tau^{#1}}\xspace}

\newcommand{\emptend}[1]{\ensuremath{{{B}}^{#1}}\xspace}
\newcommand{\aggemptend}[1]{\ensuremath{\mathbb{B}^{#1}}\xspace}

\newcommand{\varassign}[1]{\ensuremath{A_{#1}}\xspace}
\newcommand{\obsassign}[1]{\ensuremath{A_{#1}}\xspace}

\newcommand{\allvarout}{\ensuremath{Y}\xspace}
\newcommand{\varout}[1]{\ensuremath{{Y}_{#1}}\xspace}
\newcommand{\empout}[1]{\ensuremath{{\mathbb{Y}}^{#1}}\xspace}
\newcommand{\yyy}[1]{\ensuremath{{{Y}}^{#1}}\xspace}

\newcommand{\varcircm}[1]{\ensuremath{C_{#1}}\xspace}

\newcommand{\effect}[2]{\ensuremath{\mathcal{D}(#1\!,\!#2)}\xspace}

\newcommand{\empeffect}[2]{\ensuremath{\mathbb{D}(#1\!,\!#2)}\xspace}
\newcommand{\empeffectS}[2]{\ensuremath{\mathbb{D}^{1}(#1\!,\!#2)}\xspace}

\newcommand{\varselectfull}{\ensuremath{S}\xspace}

\newcommand{\obsselect}[1]{\ensuremath{S_{#1}}\xspace}

\newcommand{\shiftempeffectS}[3]{\ensuremath{\mathbb{D}^{1}(#1\!,\!#2 \!\mid\!  \obsselect{}\!=\!s)}\xspace}

\newcommand{\splitshiftempout}[3]{\ensuremath{\mathbb{Y}^{#1, \convosplit{#2}}_{#3}}\xspace}

\definecolor{lightblue}{HTML}{578de5}
\definecolor{navy}{HTML}{33368c}

\newcommand{\CorrelationalIcon}{\textcolor{gray}{$\triangle$}\xspace}
\newcommand{\CorrelationalIcons}{\textcolor{gray}{$\triangle$}s\xspace}
\newcommand{\CrossIcon}{\textcolor{lightblue}{$\square$}\xspace}
\newcommand{\CrossIcons}{\textcolor{lightblue}{$\square$}s\xspace}
\newcommand{\PairedIcon}{\textcolor{blue}{$\bigcirc$}\xspace}
\newcommand{\PairedIcons}{\textcolor{blue}{$\bigcirc$}s\xspace}

\newcommand{\convosplit}[1]{\ensuremath{{#1}}\xspace}

\newcommand{\KT}{Kendall's tau\xspace}

\newcommand{\cut}[1]{}

\definecolor{burgundy}{RGB}{138, 14, 51}

\newcommand{\rredit}[1]{{\textcolor{black}{#1}}}
\newcommand{\rrcomment}[1]{{\textcolor{black}{#1}}}
\newcommand{\rrstat}[1]{{\textcolor{black}{#1}}}

\newif\ifshowcomments
\showcommentsfalse
\newcommand{\cd}[1]{{\textcolor{blue}{#1}}}

\newcommand{\justine}[1]{{\textcolor{burgundy}{JZ: #1}}}
\newcommand{\edit}[1]{{\textcolor{navy}{#1}}}
\ifshowcomments
\else
\renewcommand{\cd}[1]{}
\renewcommand{\justine}[1]{}
\renewcommand{\edit}[1]{#1}
\fi

\begin{document}
\title{Quantifying the Causal Effects of Conversational Tendencies}
\author{Justine Zhang}
\email{jz727@cornell.edu}
\affiliation{
	\institution{Cornell University}
	\country{USA}
}

\author{Sendhil Mullainathan}
\email{sendhil@chicagobooth.edu}
\affiliation{
	\institution{Chicago Booth School of Business}
	\country{USA}
}

\author{Cristian Danescu-Niculescu-Mizil}
\email{cristian@cs.cornell.edu}
\affiliation{
	\institution{Cornell University}
	\country{USA}
}

\renewcommand{\shortauthors}{Justine Zhang et al.}



\begin{abstract}

Understanding what leads to effective conversations can aid the design of better computer-mediated communication platforms.
In particular, prior observational work has sought to identify behaviors of individuals that correlate to their conversational efficiency. 
However, translating such correlations to causal interpretations is a necessary step in using them in a prescriptive fashion to guide better designs and policies.

In this work, we formally describe the problem of drawing causal links between conversational behaviors and outcomes.  
\edit{We focus on the task of determining a particular type of policy for a text-based crisis counseling platform:
how best to allocate counselors based on their behavioral tendencies exhibited in their past conversations.}
We apply arguments derived from causal inference to underline key challenges that arise in conversational settings where randomized trials are hard to implement.
Finally, we show how to circumvent these inference challenges in our particular domain, and illustrate the potential benefits of an \edit{allocation} policy informed by the resulting prescriptive information.

\end{abstract}

\maketitle

\section{Introduction}
\label{sec:intro}
Conversations are central to the success of many consequential tasks.
Understanding how to foster more effective  discussions can guide computer-mediated platforms to better facilitate such endeavors as collaborating on large-scale projects \cite{kittur_future_2013,im_deliberation_2018,cranshaw_polymath_2011,lasecki_crowd_2012,luther_why_2010}, deliberating on law and policy \cite{fountain_prospects_2003,mcinnis_effects_2018}, 
informing and educating others \cite{yang_weakly_2015,wang_stay_2012}, 
or providing social support \cite{choudhury_language_2017,chancellor_norms_2018,saha_causal_2020}.
\edit{The growing availability of conversational data in these domains presents an opportunity to gain insights about what makes such discussions effective---a key 
step in improving these platforms.}

\edit{One promising approach towards such insights is examining how people behave in more or less effective discussions. Indeed,}
past studies have highlighted various indicators of conversational behaviors that are tied to desired downstream outcomes such as 
successful persuasion \cite{yang_persuading_2017-1,tan_winning_2016} or problem solving \cite{cranshaw_polymath_2011,niculae_conversational_2016}, or improvements in emotional state \cite{choudhury_language_2017,saha_causal_2020}. 
Conversational data was also used to characterize individuals in terms of their past conversational behaviors, providing rich signals of the roles they play in their interactions \cite{yang_who_2016,yang_persuading_2017-1} or their effectiveness in attaining conversational outcomes \cite{althoff_large-scale_2016,arguello_talk_2006,burke_introductions_2007}.
Translating these descriptive findings into \emph{prescriptive} information, however, requires determining whether the relationships between conversational behaviors and outcomes is causal in nature.

Consider, for instance, the case of a 
\edit{psychological}
crisis counseling platform---a challenging, inherently interactional domain on which
\edit{our work will focus.}
By analyzing
\edit{logs of counseling conversations,}
the platform might identify particular patterns in how counselors behave during conversations where crises are successfully addressed,
compared to during less successful interactions. 
For example, 
\edit{an analyst}
may observe that counselors who have more effective conversations also tend to exhibit more positive sentiment in their language \cite{althoff_large-scale_2016,perez-rosas_analyzing_2018}.  
Does this imply that the platform should \edit{allocate} more conversations to counselors who have the tendency to use more positive language?
Answering such prescriptive questions is a necessary prerequisite for ensuring that policies and interventions based on behavioral signals will have desired effects.
This entails probing whether observed links between behaviors and outcomes are causal in nature, a difficult task especially in settings where running randomized trials is infeasible, and where behavioral dynamics are complex.
Both of these traits are common to sensitive conversational domains such as crisis counseling.

In this work, we critically examine the task of drawing causal links between conversational behaviors and outcomes from observational data.
Our particular aim is to concretely describe the challenges of this task and highlight cases where these challenges can be addressed.
We approach this aim by way of analyzing a family of conversational settings, \textit{goal-oriented asymmetric conversational platforms}, spanning domains like customer service, mental health counseling, interviewing and tutoring.
In such settings, a platform has a dedicated roster of agents who seek to fulfill an outcome through having conversations with clients, such as an improvement in clients' mental well-being.
Crucially, the platform has some leverage in \edit{allocating} or training agents, but cannot specify the types of clients it serves.
We focus on a particular type of policy that observational data could inform, and that is highly pertinent to these settings: 
allocating agents to upcoming conversations based on their observed past behaviors. 

We first draw on the causal inference literature to provide a theoretical analysis of the inference problem involved \cite{angrist_mostly_2008,rosenbaum_design_2010}.
Using a potential outcomes framework, we articulate the causal effects we wish to estimate.
Through formalizing the inference task, we highlight two key difficulties, 
one that inherits from the broader challenge of causal inference in observational settings, and another that directly derives from the interactive nature of conversations.
This formalization also allows us to surface particular cases under which these challenges could be mitigated and to correspondingly propose solutions.

To empirically demonstrate the practical implications of our theoretical formulation,
we instantiate these inference challenges and solutions in a large dataset of counseling conversations, obtained in collaboration with a text-based crisis counseling service.
This highly consequential setting serves as a real-world example of the subclass of inference problems analyzed, 
and additionally illustrates the properties that enable these problems to be addressed.
In this context, we show that accounting for the challenges we identified allows us to make more careful inferences than naive approaches, which often overestimate the strength of the causal 
\edit{relation between conversational behaviors and outcomes.}
\rredit{We additionally probe the feasibility of an allocation policy that is based on these relations by simulating its implementation under idealized conditions, following prior work in the medical domain \cite{kleinberg_prediction_2015}.
This simulation suggests that allocating counselors based on their conversational tendencies has the potential to improve the platform's effectiveness as long as spurious relations between tendency and outcome are discounted,
and could motivate 
more realistic experimentation.}

Our theoretical and empirical analyses focus on settings like the crisis counseling service as a more tractable case study, but serve to argue a broader point:
transforming descriptive observations into prescriptive insights should be approached with care,
especially in light of the particular challenges that conversational settings present.
By explicitly articulating such challenges,
we lay the groundwork for further efforts to analyze and address causal inference problems in a wider range of conversational settings.

\section{Background and Scope}
\label{sec:background}

To clearly describe the task of making causal inferences about conversational behaviors,
we focus on formally analyzing a narrower subset of such inference problems.
In this section, we introduce and motivate our scope: 
we describe the family of conversational settings we will center our 
\edit{analyses} 
around, 
along with the particular type of policy that we would like to inform.
We also outline the causal inference challenges that we will later more rigorously take up.

\xhdr{Conversational setting}
We analyze a category of conversational settings that we term \textit{goal-oriented asymmetric conversational platforms}.
Consider a platform that maintains a roster of \textit{agents} who are expected to interact with incoming \textit{clients}.
First, the platform is \textit{goal-oriented}:
it has an overall objective that it seeks to use its agents to maximize.
Second, it is \textit{conversational} in the sense that agents work towards this objective by having conversations with clients, such that their behaviors within these conversations are consequential.
Finally, the platform has an \textit{asymmetric} degree of leverage: it can implement policies that affect its agents, but is unable to control its clients' characteristics.

This paradigm recurs across many often technologically-mediated domains like customer service~\cite{packard_im_2018,hu_read_2019}---where sales representatives interact with customers, interviews---where interviewers interact with interviewees, and education---where teachers or tutors interact with students \cite{graesser_collaborative_1995}.
As an illustrative example 
that we later revisit in more depth, consider a crisis counseling helpline.
The helpline employs a team of counselors who interact with individuals contacting the helpline in moments of mental crisis.
\edit{The overall goal of this platform}
is to help these individuals in crisis; counselors work towards this goal by having conversations with them.
The 
platform can select, train, or otherwise support its affiliated counselors.
However, it would be infeasible and even unethical to restrict the types of people who seek help from it.

\xhdr{Allocation policy}
We would like to examine how the platform can derive recommendations 
for managing its agents 
to better achieve its objective,
subject to the limited influence it has over its clients.
Here, our focus is on policies that impact how the platform \textit{allocates} its agents.
As a basic example, the platform may allocate more conversations to agents that it identifies as being more effective; as such, it may seek guidance on how best to select these effective conversationalists.
Given the inherently conversational setting, we consider policies where agents are allocated on the basis of behaviors they exhibit over past conversations they've taken,
which we refer to as behavioral \textit{tendencies}. 
As such, we analyze when these aggregate tendencies---e.g., an inclination to use more positive language or to write longer messages---can be used by the platform to identify and hence allocate conversations to more effective agents.
Intuitively, observing that certain behavioral signals are correlated with desired conversational outcomes would suggest that the platform should use tendencies 
\edit{inferred from}
these signals to allocate agents.
The remainder of our work more rigorously examines this intuition.

We note that using behavioral tendencies to allocate agents is one of many policies that a platform could pursue. 
Here, we briefly outline some alternatives and motivate our particular focus.
First, the platform may wish to allocate agents without accounting for their past behaviors---for
instance, 
it may instead rely on past performance, or on demographic and personality attributes \cite{lykourentzou_personality_2016-1,kim_what_2017,bear_role_2011,woolley_what_2011,baer_personality_2008}.
However, as noted in the introduction, past work illustrates that conversational behaviors can provide rich signals of a conversation's outcome or an agent's characteristics;
here, we specifically take up the potential usefulness of these signals 
in guiding concrete policies. 
We later empirically compare the effectiveness of conversational signals to these non-conversational attributes.

Second, more sophisticated allocation policies could extend the demonstrative approach considered 
here---\edit{for instance, the platform could}
match particular agents with conversations they are particularly well-suited for \cite{maki_roles_2017}. 
We leave an analysis of such policies to future work, noting that these more informed allocations require additional information (such as the nature of the client involved) which may not 
\edit{be readily}
available at the start of a conversation. 

Finally, we contrast a policy of allocating agents with one that \textit{trains} agents to adopt particular behaviors  \cite{demasi_towards_2019,schwalbe_sustaining_2014,andrade_avatar-mediated_2010}.
Both allocation and training-based policies could be informed by inferences about how behaviors and outcomes relate. 
We later revisit the training approach 
\edit{in the discussion (Section \ref{sec:discussion})} 
to suggest that it shares the inference challenges that the allocation policy is subject to, but comes with additional difficulties as well, which we leave for future work.

\xhdr{Overview of inference challenges}
As a precondition to implementing any allocation policy,
the platform would need to ensure that the policy could actually have a desired effect.
Concretely, we must consider a counterfactual question: if the platform had allocated another agent with a different tendency to a conversation, would the conversation have had a better outcome?
While this question could in principle be addressed via randomized experiments, 
an experimental approach is often infeasible given the sensitivity of a conversational setting like counseling, and the difficulty of specifying treatments involving complex linguistic or interactional signals \cite{egami_how_2018,wang_when_2019}.
Addressing such inherently counterfactual questions with observational data has been a core focus of causal inference (for surveys, see \cite{angrist_mostly_2008} and \cite{rosenbaum_design_2010}). 
Such literature, however, has not dealt with the setting of conversations, which presents additional challenges that we identify and address in this work.

To outline the 
difficulties
of the inference task in a conversational domain, consider a naive approach 
for
relating conversational behaviors and outcomes:
if we observe that good outcomes follow conversations where agents exhibit a certain behavior, we may naively infer that this behavior is a useful signal of effectiveness.
For instance, suppose we find that client mood tends to improve after conversations involving agents who use language with a greater degree of positive sentiment.
Such a finding could motivate us to allocate more positive agents to more future conversations.

At a high level, this initial approach suffers from a crucial pitfall:
while an outcome may indeed arise as a result of an agent's 
\edit{behaviors,}
many \textit{circumstantial} factors could also influence both the outcome and the nature of the conversation, and thus the \edit{behaviors} that the agent exhibits.
As such, our observations of the relation between a behavior and an outcome are confounded with circumstances of the conversation that neither the agents nor the platform can influence.
For instance, an agent may say more positive things in a circumstance involving a congenial client
who might also be more easily satisfied.
However, in a situation involving a client with a genuinely 
\edit{difficult}
situation, a tendency for positivity may not even be appropriate, let alone effective.
This means that a naive correlational approach cannot answer the counterfactual question posed above; 
in particular, 
\edit{the approach cannot}
inform us on how more positive agents would fare in conversations with less congenial clients.

\section{Formulating the inference task}
\label{sec:formulation}

We now proceed to more rigorously examine the entanglement between behavior, outcome and circumstance, 
focusing on the policy of allocating agents given their conversational tendencies.
While the ideas we subsequently discuss are broadly relevant to other policies (such as those for training agents), 
we can more tractably address the inference task in the allocation policy,
and hence present a formal analysis of this policy as an illustrative starting point.

In particular, the allocation policy takes an aggregated view:
the platform makes allocations based on how agents tend to behave over their past conversations.
Intuitively, taking agent-level aggregates decouples our analyses from the circumstances of any one interaction:
while an agent's behavior in a single conversation may be constrained by circumstantial peculiarities,
over many conversations, their personal inclinations may materialize as conversational tendencies. 
Likewise, an agent may exhibit a systematic \textit{propensity} to elicit certain outcomes, even if the outcome of a single interaction is contingent on its circumstance. 

In what follows, we draw on the causal inference literature to formally examine the inference task underlying the allocation policy \cite{angrist_mostly_2008,rosenbaum_design_2010}.
First, we define this task in terms of the causal effect of allocation that we wish to estimate.
We then discuss the challenges we face in quantifying this effect.
We decompose these challenges into two key difficulties that stem from the \textit{observational} nature of our data
and the \textit{interactional} nature of our conversational setting.
We analyze each of these challenges by concretely identifying biases that arise in naive estimators of the effect of allocation, and then describe 
\edit{particular cases} 
under which these biases can be addressed.

\xhdr{Inference task: estimating the allocation effect}
Our goal is to evaluate the potential effectiveness of a policy that allocates agents to conversations, given their conversational tendencies.
We now discuss the central measurement in this task, which corresponds to the counterfactual question introduced in the preceding section:
given two agents \inter{J} and \inter{K}, who have different tendencies with respect to some behavioral signal 
(e.g., \inter{J} tends to use more positive language than \inter{K}),
\rrcomment{what is the effect}
of allocating one agent to a conversation versus the other, on a given outcome?
We henceforth refer to this quantity as the \textit{allocation effect}.

Under our observational approach, we wish to estimate the allocation effect from data on conversations that \inter{J} and \inter{K} have already taken.
As such, we must use the data in two ways.
First, we must 
\edit{use past observations to estimate}
the propensity of each agent to get a desired outcome
(e.g., proportion of their clients who improved their mood).
Second, we must estimate each agent's behavioral tendencies from their past conversations.\footnote{Indeed, conversational data seldom comes with a priori labels of how agents tend to act; we may contrast this data-driven approach with self-reported indicators.}

\begin{figure}[t!]
\centering
\includegraphics[width=0.25\textwidth]{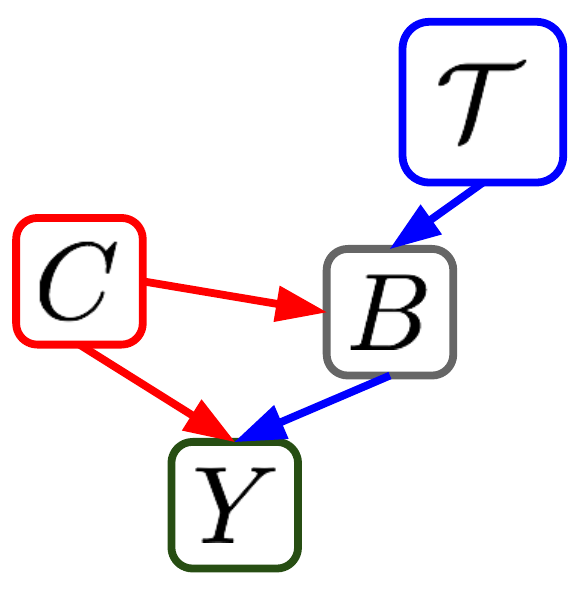}
\caption{Graphical representations of the key dependencies underlying the inference task, between tendency \vartend{}, outcome \varout{}, behavior \emptend{} and circumstance \varcircm{}. Our goal is to estimate the effect of tendency on outcomes (blue path), however the circumstances under which the behaviors and outcomes are observed confound this estimation (red arrows).
}
\Description{Depiction of dependencies between tendency, outcome, behavior and circumstance.}
\label{fig:basic_deps}
\end{figure}

In order for our estimate of the allocation effect to have a causal interpretation, 
we must ensure it can be directly ascribed to differences in the tendencies of \inter{J} and \inter{K},
rather than to differences in the circumstances of the conversations in which \inter{J} and \inter{K}'s outcomes and behaviors were observed.
As noted in the preceding section, these conversational circumstances can shape both the outcome of a conversation and an agent's behavior within the conversation, which thus become 
\edit{entangled.}
These problematic dependencies are summarized in the 
graphical representation~\cite{pearl_causal_1995} depicted in Figure~\ref{fig:basic_deps}.
\edit{We would like to estimate the effect of (allocating) tendencies \vartend{} on outcomes \varout{} (blue path);
to this end, we must use behaviors \emptend{} and outcomes \varout{} observed under particular circumstances \varcircm{}.
These circumstances can determine both behaviors and outcomes (red paths);
our challenge is thus to somehow disentangle the effects of circumstances and tendencies.}

\xhdr{Potential outcomes formulation}
To formally highlight the biases that are incurred as a result of this entanglement, we mathematically express the allocation effect in terms of the potential outcomes framework \cite{angrist_mostly_2008,rosenbaum_design_2010}.
Let \vartend{} be a random variable denoting a conversational tendency of agents, 
and suppose that agents \inter{J} and \inter{K} have different tendencies \obstend{\inter{J}} and \obstend{\inter{K}}.
Let \allvarout{} be a random variable denoting a conversational outcome.
The allocation effect is then the expected difference in outcome if \inter{J}, rather than \inter{K}, is allocated to a conversation:

\begin{align}
 \effect{\obstend{\inter{J}}}{\obstend{\inter{K}}} 
 \!=\! \condexpt{\allvarout}{\vartend\!=\!\obstend{\inter{J}}}\! -\! \condexpt{\allvarout}{\vartend\!=\!\obstend{\inter{K}}} \label{eq:basic_diff}
\end{align}

Let \empeffect{\obstend{\inter{J}}}{\obstend{\inter{K}}} denote an estimate of \effect{\obstend{\inter{J}}}{\obstend{\inter{K}}} from the data.
Formally, this estimate has a causal interpretation if it is unbiased, i.e., $\expt{\empeffect{\obstend{\inter{J}}}{\obstend{\inter{K}}}}\!\!=\!\!\effect{\obstend{\inter{J}}}{\obstend{\inter{K}}}$.
Conversely, the estimate fails if it is contingent on the circumstances \varcircm{} under which the observed conversations occurred.\footnote{Throughout, the notation we use adopts the following convention: 
uppercase denotes random variables (e.g., \allvarout{}, \vartend{} and $\mathcal{D}$ are random variables for conversational outcome and tendency, respectively),
lowercase denotes realizations of these variables (e.g., \obstend{\inter{J}} is an observed value of \vartend{}),
and empirical estimators are listed in blackboard bold (e.g., $\mathbb{D}$ is an empirical estimate of $\mathcal{D}$ based on the observed data).}

As we have intuitively noted and as shown in Figure \ref{fig:basic_deps}, such dependencies on \varcircm{} arise when we estimate \allvarout{} with observed outcomes,
and \vartend{} with observed behaviors.
We now proceed to articulate the challenges that are incurred from these dependencies.
For each challenge, we provide an intuitive description supplemented with a graphical representation of the relationships between the variables involved \cite{pearl_causal_1995}, before drawing on potential outcomes arguments to formally express the corresponding biases \cite{angrist_mostly_2008,rosenbaum_design_2010}.
Our formal descriptions also point to particular settings with properties that enable us to mitigate these biases,
\edit{and we discuss  solutions that make use of these properties as well.}

\subsection{Estimating outcomes: bias from observed assignment}
\label{sec:formulation_observation}

We first address the difficulties stemming from estimating agents' propensities for outcomes \allvarout{} using our observations of their past conversations. 
To simplify the discussion, we provisionally suppose that we are given explicit labels of the agents' tendencies, returning to this point in the next subsection (\ref{sec:formulation_interaction}).

At a high level, our measurement of how tendencies relate to outcomes suffers from a problem that pervades observational studies: 
we can only observe outcomes in conversations that were actually \textit{assigned} to agents exhibiting these tendencies.
Here, we describe this problem in the context of conversations.

Let \obsassign{} denote the observed assignment---i.e., the matching between each agent \inter{J} and their conversations in the data.
The assignment mechanism potentially exposes different agents to contrasting circumstances:
for example, agent \inter{J} may be assigned to more challenging clients than \inter{K}.
As such, these assignment-induced differences in circumstance,
rather than differences in the agents' tendencies,
could drive observed differences in outcome.
\edit{In this way, \obsassign{} skews our estimation of the allocation effect.}

\begin{figure}[t!]
\centering
\includegraphics[width=0.75\textwidth]{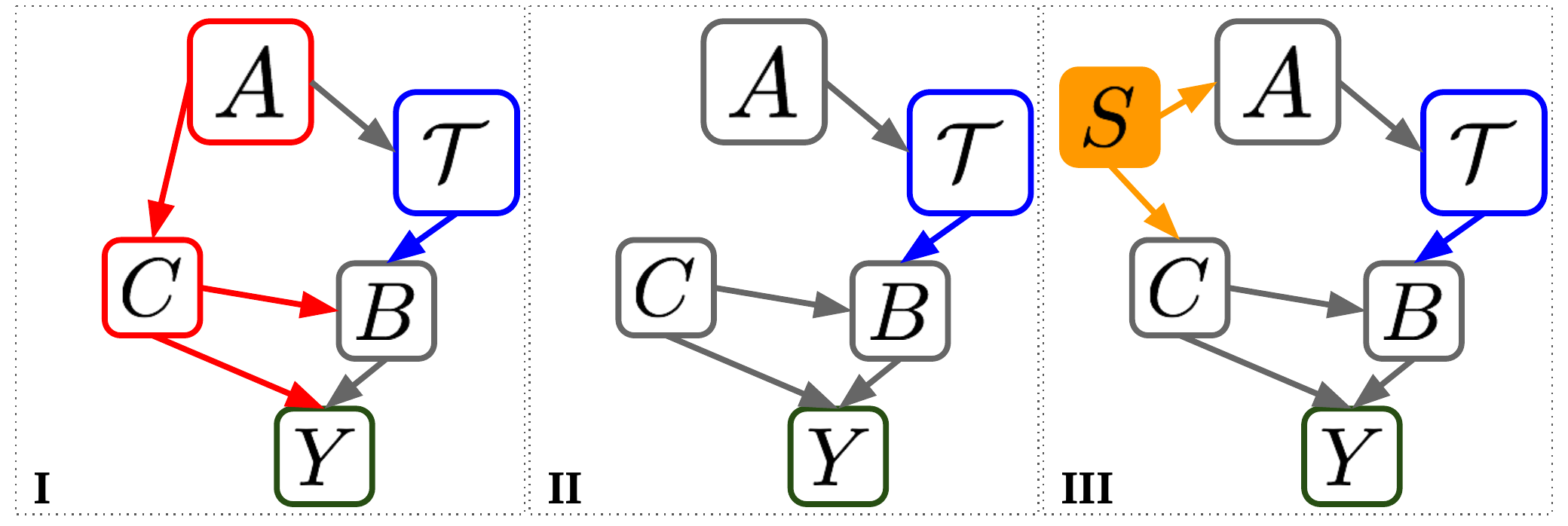}
\caption{Graphical representations of the dependence between assignment \varassign{} and outcome \varout{} through behavior \emptend{} and circumstance \varcircm{} that result from the observational nature of our analyses, giving rise to the selection bias exposed in~\eqref{eq:assignment_circm}. \textbf{I}:~the problematic pathways from \varassign{} to \varout{}; \textbf{II}:~an idealized setting where conversations are randomly assigned to agents, in which the dependency is trivially broken; \textbf{III}:~a scenario
where assignment is governed by a set of observable \textit{selection variables} \obsselect{}. 
}
\Description{Graphical representations of the dependence between assignment and outcome through behavior and circumstance, giving rise to bias from observed assignment.}
\label{fig:assignment_bias}
\end{figure}

The graphical model depicted in Figure \ref{fig:assignment_bias} highlights the problematic dependencies between tendency \vartend and outcome \varout{}, as indicated by the red edges:
assignment \obsassign{} determines both \vartend and \varcircm{}, which  in turn determine \varout{}. 
As such, we cannot discount the effect of differences in assignment (red), beyond differences in tendency (blue), on the observed outcome.

\xhdr{Potential outcomes formulation}
To surface the bias incurred from assignment, we formally examine the estimation of \allvarout{}.
As a first attempt, we can estimate the propensity of an agent \inter{J} to get an outcome using the average outcome over their past conversations, denoted \empout{\inter{J}}.
As such, we would measure \effect{\obstend{\inter{J}}}{\obstend{\inter{K}}} 
as $\empeffect{\obstend{\inter{J}}}{\obstend{\inter{K}}} \!=\!\empout{\inter{J}} \!-\! \empout{\inter{K}}$.

We note that our empirical estimators are contingent on \obsassign{}, i.e., we can only observe \inter{J} in the conversations in which they actually participated.
\edit{As such,}

$$\expt{\empout{\inter{J}}} = \condexpt{\varout{}}{\vartend=\obstend{\inter{J}}, \obsassign{}=\inter{J}}$$ 

\edit{Substituting this expression into the above equation for \empeffect{\obstend{\inter{J}}}{\obstend{\inter{K}}}, 
we see that our estimator of the allocation effect is biased:\footnote{In the last derivation we subtract and re-add the second term. }}

\begin{align}
\expt{\empeffect{\obstend{\inter{J}}}{\obstend{\inter{K}}}} & =  \expt{\empout{\inter{J}}\!-\! \empout{\inter{K}}} \nonumber\\
 &\!=\! \condexpt{\varout{}}{\vartend\!=\!\obstend{\inter{J}}\!, \!\obsassign{}\!=\!\inter{J}}  
 \!-\! \condexpt{\varout{}}{\vartend\!=\!\obstend{\inter{K}}\!,\! \obsassign{}\!=\!\inter{K}} \nonumber\\
& \!=\! \condexpt{\varout{}}{\vartend\!=\!\obstend{\inter{J}}\! , \! \obsassign{}\!=\!\inter{J}} \!-\! \condexpt{\varout{}}{\vartend\!=\!\obstend{\inter{K}}\! , \! \obsassign{}\!=\!\inter{J}} \label{eq:assignment_tend}\\
& \!+\! \condexpt{\varout{}}{\vartend\!=\!\obstend{\inter{K}}\! , \! \obsassign{}\!=\!\inter{J}} \!-\! \condexpt{\varout{}}{\vartend\!=\!\obstend{\inter{K}}\! , \! \obsassign{}\!=\!\inter{K}}\!\label{eq:assignment_circm}
\end{align}

\noindent The equations highlight that our observed difference could have two sources. 
The first \eqref{eq:assignment_tend} corresponds to the effect of varying the tendencies over a shared set of circumstances (i.e., that were assigned to \inter{J}).
This is the value we need to estimate in order answer the counterfactual question:
what outcomes would have been attained had the conversations that were assigned to \inter{J} been instead handled by an agent with a different tendency \obstend{\inter{K}}?
The second \eqref{eq:assignment_circm} reflects the \textit{selection bias} that arises because \inter{J} and \inter{K} were actually exposed to different circumstances via assignment,
as illustrated in Figure \ref{fig:assignment_bias}I.

\xhdr{An idealized setting: random assignment}
As with many causal inference questions, selection bias would be eliminated if agents were \textit{randomly assigned} to conversations, and are hence exposed to the same distributions of circumstances.
As such, observed differences in outcome could no longer be ascribed to assignment-induced differences in circumstance.
Formally, random assignment makes assignment and outcome independent for each agent (Figure \ref{fig:assignment_bias}II),
such that the problematic term \eqref{eq:assignment_circm} trivially cancels out.

However, this selection bias remains in more realistic conversational settings,
\edit{where assignment mechanisms are seldom random.}
In the extreme, if an agent \textit{selects} their conversations, a record of positive conversational outcomes could be ascribed to picking clients who are easier to help,
rather than having some replicable conversational proficiency.
The problem persists beyond self-selection---e.g., agents who work during the day may encounter more congenial clients than those who work at night. 

\xhdr{A limited solution: controlling for circumstance}
We may try to mitigate selection biases by controlling for the conversational circumstances \varcircm{}, 
for instance by
comparing \empout{\inter{J}} and \empout{\inter{K}} only over conversations that match on attributes of the circumstance, 
e.g., are about the same issue.
Indeed, many prior studies of conversations have employed 
such techniques
\rrcomment{\cite{tan_winning_2016,zhang_conversations_2018,sridhar_estimating_2019,jaech_talking_2015,pavalanathan_emoticons_2015,choudhury_language_2017,saha_causal_2020}}.
\textit{Completely} controlling for circumstance certainly breaks the problematic pathway from \obsassign{} to \varout{}:
Figure \ref{fig:assignment_bias}I shows that the two variables are conditionally independent given \varcircm{} and \vartend (formally written as $\varout{}\perp\!\!\!\perp\obsassign{}  \ | \{\varcircm{},\vartend\}$).\footnote{Conditional independence corresponds to the criterion of d-separation in the graphical representation
\cite{pearl_causal_1995}.}

However, this approach is fundamentally limited: 
we can only control for the circumstantial attributes that we can observe.
This leaves other important but inaccessible aspects 
(e.g., the client's mental state) 
unaccounted for.

\xhdr{Tractable setting: observed selection variables}
We now describe a subset of settings under which this bias can be mitigated, involving assignment mechanisms with some additional structure.
In particular, suppose that assignment is random up to a set of completely observable \textit{assignment selection variables} \obsselect{}
(Figure \ref{fig:assignment_bias}III, orange edges).
As a natural example, consider conversational platforms where agents work during different \textit{shifts}, and clients are randomly assigned to agents within each shift time.
While different agents and clients may select different shifts,
within a single shift these factors play no role in who gets assigned to whom;
furthermore, for each conversation, the platform knows the shift in which it took place. 
Beyond shift times, other examples of selection variables include geographic location and organizational divisions like departments of a store.\footnote{We are effectively using the assignment of agents as valid instrument, conditional on shift, for the kinds of conversation the client is exposed to \cite{angrist_credibility_2010,brito_generalized_2012,pearl_testability_2013}.}

Importantly, conditioning on \obsselect{} breaks the pathway between \obsassign{} and \varout{}; 
that is, \varout{} and \obsassign{} are {conditionally independent} given \obsselect{} and \vartend{} ($\varout{}\perp\!\!\!\perp\obsassign{}  \ | \{\obsselect{} ,\vartend\}$). 
Controlling for selection variables can be seen as a special case of controlling for observable circumstantial attributes, 
where we know how these attributes \obsselect{} are related to the assignment mechanism.
\textit{Within each value of the selection variable},
our observations of agents' conversational outcomes are hence decoupled from circumstantial differences due to assignment.
As such, we modify our estimator to first measure the allocation effect for a particular selection variable (e.g., within a shift),
comparing outcomes attained by agents with tendencies \obstend{\inter{J}} versus \obstend{\inter{K}}
only for conversations with that selection variable.

Formally, for a given selection variable $s$, denote the corresponding estimator of the allocation effect as \empeffect{\obstend{\inter{J}}}{\obstend{\inter{K}}| \obsselect{}\!=\!s}.
By conditional independence, we have that:
\begin{align}
& \condexpt{\varout{}}{\vartend=\obstend{\inter{J}}, \obsassign{}=\inter{J}, \obsselect{}=s} \nonumber\\
& = \condexpt{\varout{}}{\vartend=\obstend{\inter{J}}, \obsassign{}=\inter{K}, \obsselect{}=s} \nonumber \\
& = \condexpt{\varout{}}{\vartend=\obstend{\inter{J}},  \obsselect{}=s} \nonumber
\end{align}
Thus, after conditioning on \obsselect{}, the bias \eqref{eq:assignment_circm} cancels out. 
That is, 
among conversations with the same \obsselect{}, 
empirical differences in outcome are entirely driven by tendency: 
\begin{align}
 & \expt{\empeffect{\obstend{\inter{J}}}{\obstend{\inter{K}}| \obsselect{}\!=\!s}} \!=\! \nonumber\\
 &\!=\! \condexpt{\varout{}}{\vartend\!=\!\obstend{\inter{J}},  \obsselect{}\!=\!s} \!-\! \condexpt{\varout{}}{\vartend\!=\!\obstend{\inter{K}},\obsselect{}\!=\!s}\label{eq:cond_on_shift}
 \end{align}

Repeating this matching process across all \obsselect{} then yields an aggregate measurement of outcome differences arising from varied tendencies, rather than from differences in assignment.

\subsection{Estimating tendencies: bias from interactional effects}
\label{sec:formulation_interaction}

We now address the difficulties stemming from estimating agents' tendencies \vartend using our observations of their past behaviors.
To simplify the discussion, we suppose that the difficulty in estimating outcomes, as described in the preceding section (\ref{sec:formulation_observation}), has been fully addressed.

At a high level, the problem we face stems from the interactional nature of conversations:
the behavior of an agent both shapes, and is constantly shaped by the behavior of the other participant. 
As such, our measurement of an agent's tendencies, and hence our inferences about the relation between tendency and outcome, is skewed by the circumstances that agents inevitably react to in conversations.
At an extreme, we may observe that agents say ``you're welcome'' precisely after clients thank them.
This does not necessarily mean that saying ``you're welcome'' is a behavioral inclination some agents have, beyond a reaction to the preceding interaction;
it certainly does not follow that we should encourage more frequently saying ``you're welcome''.

\begin{figure}[t!]
\centering
\includegraphics[width=0.75\textwidth]{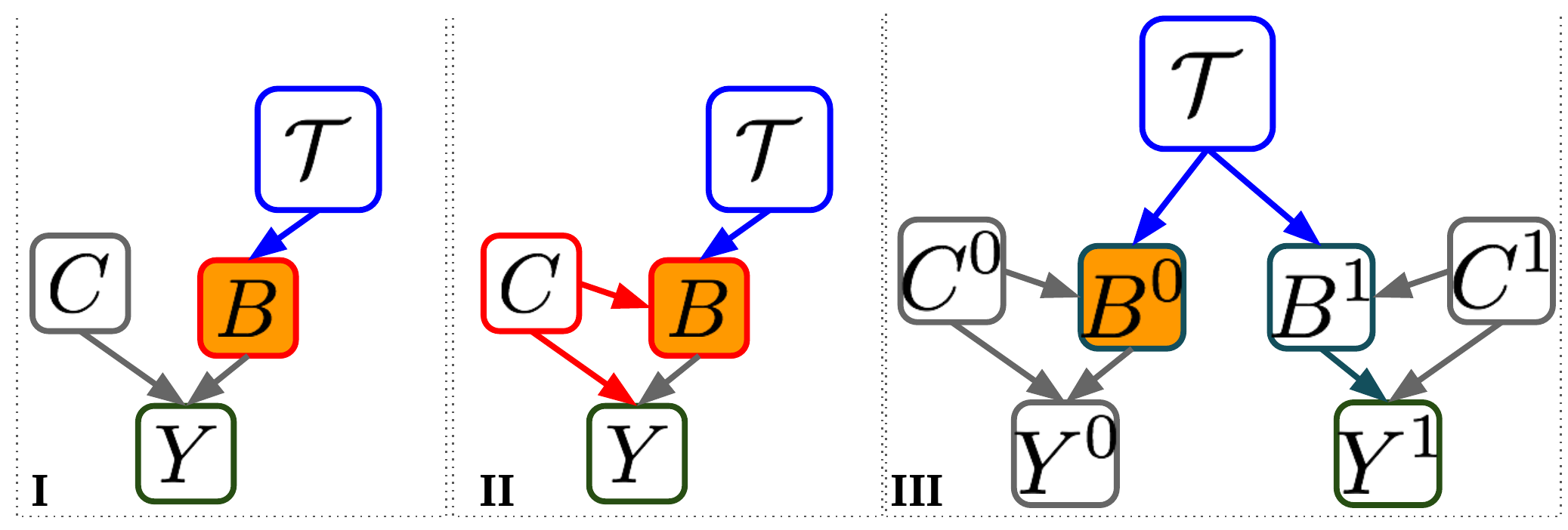}
\caption{Graphical representations of the entanglement between circumstances \varcircm{}, behaviors \emptend{} and outcomes \varout{}{}, giving rise to the bias in  \eqref{eq:circularity_circm}. \textbf{I}: dependencies in a non-interactional setting; \textbf{II}: problematic dependencies
when \emptend{} interacts with circumstances \varcircm{} that also shape \varout{}{}; \textbf{III}: our approach, 
observing behaviors and outcomes on different splits of data.}
\Description{Graphical representations of the entanglement between circumstances, behaviors and outcomes, giving rise to bias from interactional effects.}
\label{fig:circularity}
\end{figure}

\xhdr{An interactional problem} 
As a thought experiment, consider a \textit{non-interactional} domain where an agent's behavior can affect an outcome without any interaction with the client---a ``secret santa'' paradigm where an agent, the gift-giver, has no back and forth with their recipient 
(Figure \ref{fig:circularity}I). 
In this case an agent's behavior is purely a reflection of the agent's tendencies (e.g., an inclination for cheap gifts);
and an empirical mismatch between \vartend{} and \emptend{} simply reflects the noise with which a tendency gives rise to a behavior. 
As we accrue more observations of the agent, we would expect such mismatches to diminish.

In contrast, in an interactional setting, these factors are problematically entangled (Figure \ref{fig:circularity}II, red path): 
since the agent inevitably reacts to the client's behavior,
\emptend{} reflects  \varcircm{} as well as \vartend{}. 
Furthermore,  \varcircm{}  can impact outcomes \allvarout. 
\edit{An agent's observed behavior hence constrains the distribution of \varcircm{} that could have yielded our observed outcomes;
as with nonrandom assignment, differences in observed outcomes once again could reflect differences in circumstance as well as in tendency.}

\xhdr{Potential outcomes formulation}
Formally, let \emptend{} be a random variable denoting observed agent behaviors.
We use an aggregate of \inter{J}'s past behaviors, denoted 
\rrcomment{\aggemptend{\inter{J}}},
to measure \obstend{\inter{J}}. 
Our empirical estimators are hence contingent on these observed behaviors:

\rrcomment{$$\expt{\empout{\inter{J}}} = \condexpt{\varout{}}{\vartend=\obstend{\inter{J}}, \emptend{}=\aggemptend{\inter{J}}}$$}

Again, we highlight the bias in estimator \empeffect{\obstend{\inter{J}}}{\obstend{\inter{K}}}:

\rrcomment{\begin{align}
& \expt{\empout{\inter{J}}\!-\! \empout{\inter{K}}} \nonumber\\
& \!=\! \condexpt{\varout{}}{\vartend\!=\!\obstend{\inter{J}}\!,\! \emptend{}\!=\!\aggemptend{\inter{J}}}\!-\!\condexpt{\varout{}}{\vartend\!=\!\obstend{\inter{K}}\!,\! \emptend{}\!=\!\aggemptend{\inter{K}}}\nonumber \\
&\!=\! \condexpt{\varout{}}{\vartend\!=\!\obstend{\inter{J}}\!,\! \emptend{}\!=\!\aggemptend{\inter{J}}}\!-\!\condexpt{\varout{}}{\vartend\!=\!\obstend{\inter{K}}\!,\! \emptend{}\!=\!\aggemptend{\inter{J}}} \!\label{eq:circularity_tend} \\
&\!+\! \condexpt{\varout{}}{\vartend\!=\!\obstend{\inter{K}}\!,\! \emptend{}\!=\!\aggemptend{\inter{J}}}\!-\!\condexpt{\varout{}}{\vartend\!=\!\obstend{\inter{K}}\!,\! \emptend{}\!=\!\aggemptend{\inter{K}}} \!\label{eq:circularity_circm}
\end{align}}

As before, two factors contribute to the observed difference in outcome. 
The first \eqref{eq:circularity_tend} arises from a difference in tendencies. 
The second \eqref{eq:circularity_circm}, as we've described above and as depicted in Figure \ref{fig:circularity}II, reflects a difference in circumstances, and is inherent to the interactional nature of conversations.

\xhdr{A limited solution: ignoring the interaction}
The factor in \eqref{eq:circularity_circm} intuitively compounds as the conversation progresses and an agent's behavior becomes increasingly contingent on the circumstances.
As such, we may seek to dampen this bias by only considering behaviors from the start of the conversation, before behavior and circumstance become tightly coupled.
Indeed, prior work has taken this limited view of conversations \cite{althoff_large-scale_2016,zhang_conversations_2018} with this confound in mind.
However, insofar as this approach does not fully address the bias incurred by interaction, it also constrains the scope of the conversational tendencies we can study. 

\xhdr{Tractable setting: separable sets of conversations} 
To factor out this interactional bias,
we must decouple our observations of agent behaviors and outcomes from the conversational circumstances they are both tied to.
We consider a simple fix: 
\rredit{
for each agent, we measure their behaviors over a \textit{subset} of the conversations they've taken,
and use a \textit{separate} set of conversations to measure the outcomes they elicit.}\footnote{While we solve a different problem, our solution is analogous to 
separating train and test sets to mitigate overfitting \cite{egami_how_2018}---here we ``train'' our measurements of tendencies and ``test'' their effects on separate data splits. \rredit{Note that throughout, we use ``subset'' to refer to a collection of conversations, not to a subset of messages within a single conversation.}}

Formally, suppose we split each of \emptend{}, \varout{}, \varcircm{} into two random variables, one for each subset.
As shown in Figure \ref{fig:circularity}III, the only pathway connecting an agent's behaviors and outcomes \textit{across} these splits is via their conversational tendencies. 
That is, \emptend{0} and \yyy{1} are conditionally independent given \vartend{} (i.e., $\yyy{1}\perp\!\!\!\perp\emptend{0}  \ | \vartend$), so

\rrcomment{\begin{align}
&\condexpt{\yyy{1}}{\vartend=\obstend{\inter{J}}, \emptend{0}=
\aggemptend{\inter{J},0}}=\condexpt{\yyy{1}}{\vartend=\obstend{\inter{J}}}\nonumber
\end{align}}

\noindent and the bias term \eqref{eq:circularity_circm} cancels out.

Such a solution is applicable to conversational platforms where agents take many conversations, 
and where different subsets of these conversations are separable from each other,
as in a common scenario where clients contact a platform for ad-hoc purposes.
We may contrast these conditions with settings in which clients influence each other or recur across multiple interactions.

\section{Empirical demonstration}
\label{sec:demonstration}

Having developed a general description of our inference task and 
\rrcomment{its challenges,}
we now demonstrate these ideas empirically.
In particular, we consider a real-world example of an asymmetric conversation platform: 
a large-scale crisis counseling service.

In what follows, we introduce this setting, describe the particular dataset we examine, 
and explain how it is illustrative of our theoretical formulation. 
We then study the allocation effect of a few simple tendencies,  
comparing naive estimators to approaches informed by our preceding analyses.
\rredit{Finally, we estimate the effects of a policy of allocating counselors via a simulated experiment, showing}
how a careful consideration of conversational tendencies can provide \rrcomment{an informative} starting point in evaluating this policy.

\subsection{\rrcomment{Setting: Crisis counseling conversations}}
\label{sec:demonstration_data}

The crisis counseling platform provides a free 24/7 service where \textit{counselors}---playing the role of agents---have 
conversations  via text message with clients in mental distress who contact the platform, henceforth \textit{texters}.
We accessed the complete collection of anonymized conversations in collaboration with the platform, Crisis Text Line,\footnote{Access to the data is by application, at \footnotesize{\url{https://www.crisistextline.org/data-philosophy/research-fellows/}}. The extensive ethical considerations, and policies accordingly implemented by the platform, are detailed in \citet{pisani_protecting_2019}.} and with IRB approval; counselors and texters have consented to make their data available for research purposes.

The counseling platform is a particularly consequential example of a goal-oriented asymmetric conversational platform.
The platform's overall goal is to better support texters through their distress;
counselors aim at this objective in each conversation they take with a texter.
These conversations are challenging and complex, and are typically quite substantial, averaging 26 messages long.
They give rise to a rich array of conversational behaviors and interactional dynamics which may impact a texter's experience \cite{althoff_large-scale_2016,zhang_balancing_2020,zhang_finding_2019}.

\xhdr{Conversational clients: texters}
Texters contact the platform with a variety of issues, ranging from depression to work problems to suicidal ideation. 
They encompass a broad range of demographic and geographical attributes; 
seasonal changes 
and current events
can also shape the types of texters who contact the service. 
Crucially, the service is open for anyone to reach out to at any time.
Conversations with these texters thus span a challenging diversity of circumstances.

\xhdr{Platform agents: counselors} 
Counselors with the service are dedicated volunteers who are selected and trained by the platform.
The long-term nature of counselors' engagement with the platform
underlines the practical 
\rrcomment{relevance}
of an agent-level allocation policy.
In taking a long-term view of counselor behavior, we focus on analyzing the subpopulation of 4,861 counselors who take at least 80 conversations.
These counselors constitute 34\% of the total population, and have taken a total of 1,180,473 conversations, or 83\% of conversations on the platform to date.
\rredit{Henceforth, all of the statistics we report are computed over the first 80 conversations taken by each of these sufficiently prolific counselors.}

\xhdr{Conversational outcomes}
For the purposes of our present demonstration, we consider two complementary signals of a conversation's outcome 
\rredit{that are used by the platform to assess the conversations that take place on it}:

\begin{itemize}
\item \textbf{Texter rating}: 
After each conversation, texters are surveyed on whether or not the interaction was \textit{helpful}. 
Out of \rrstat{29\%} 
of conversations that receive ratings, 87\% are positively rated. 
\rredit{Prior computational analyses of counseling conversations have also used such survey responses as an indicator of conversation quality \cite{althoff_large-scale_2016,zhang_balancing_2020}.}
\item \textbf{Conversational closure}: 
Ideally, conversations \textit{close} at a moment that feels appropriate for both the counselor and texter. 
However, not all texters remain engaged in a conversation.
A counselor who is faced with an unresponsive texter ends the conversation after following a standardized protocol specified by the platform.\footnote{In particular, this standardization minimizes the impact of the counselor's inclinations in determining this  outcome.}  
\rrstat{72\%}
of conversations are properly \textit{closed} in this sense, while the rest end in texter disengagement. 
\end{itemize}

\rredit{In general, evaluating the success of a counseling conversation is difficult \cite{tracey_expertise_2014,hill_clientcentered_2000}. 
For instance, eliciting feedback from texters is challenging, as evidenced by the relatively low proportion of post-conversation surveys with responses;
the ratings obtained may also reflect a biased sample of texters who decide to fill out the survey.\xspace
\footnote{\rredit{We note that that there is an insignificant correlation (\KT$=0.02$) between the propensity of a counselor to receive ratings, and their propensity for positive ratings (among their rated conversations), suggesting that at the counselor level, analyses focusing on counselors for whom enough ratings are observed would not be skewed towards counselors whose conversations are better- or worse-perceived.}}
On the other hand, while the closure outcome is well-defined over all conversations, it is a less precise indicator---texters may disengage from a conversation for numerous reasons which may be extrinsic to the interaction, such as a low phone battery.\footnote{\rredit{The potential for several factors beyond what's recorded in the data to relate to closure exemplifies the limited efficacy of controlling for observable attributes of the circumstance.}}
Given the focus of our work on rigorously probing potential causal relations between behavioral tendencies and outcomes, we leave the problem of developing richer and more reliable measures of conversational outcomes to other work.
We also note that alternative outcomes would be still be subject to the circumstantial confounds detailed in our general description in Section \ref{sec:formulation}.}

\begin{figure}[t]
\centering
\includegraphics[width=0.7\textwidth]{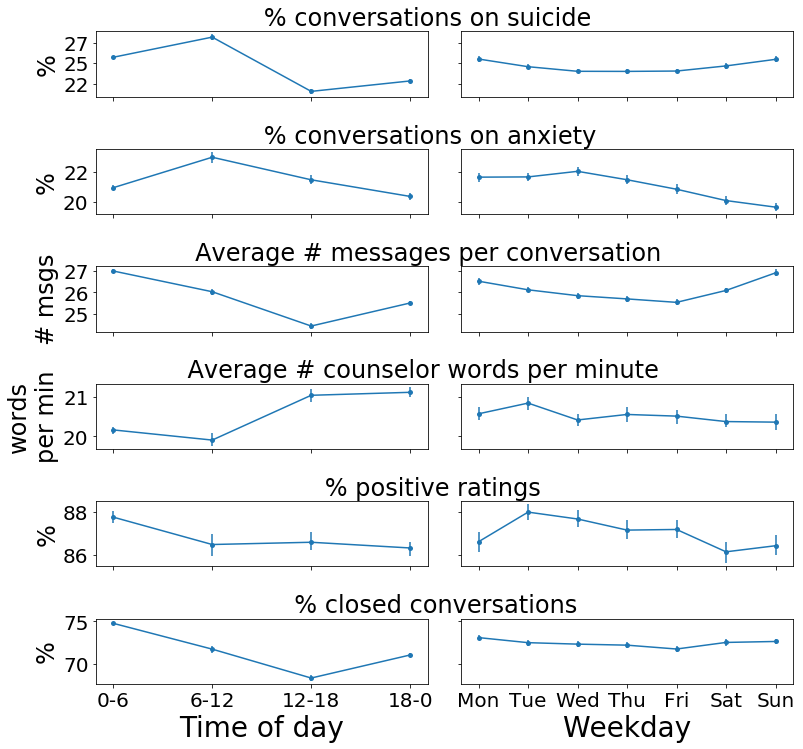}
\caption{\rredit{Issue frequency, average conversation length, counselor speed, and outcomes across a day or week. Error bars represent bootstrapped 95\% confidence intervals.}
}
\Description{Issue frequency, average conversation length, counselor speed, and outcomes across a day or week.}
\label{fig:shift_time_variation}
\end{figure}

\xhdr{Circumstance-based confounds}
Intuitively, both outcomes are heavily dependent on the texter a counselor interacts with, and the context in which a conversation takes place.
\rredit{This underlines the need to address circumstantial factors when relating behavioral tendencies to outcomes.}
As a concrete demonstration of the salience of these circumstantial factors, we observe that the time at which a conversation takes place is related to the types of issues a texter experiences, a counselor's conversational behaviors, and the outcomes that result, as shown in Figure \ref{fig:shift_time_variation}.
\rrstat{For instance,
the share of conversations involving suicidal ideation peaks in the morning and drops in the afternoon (28\% vs. 22\% of conversations);
conversations are especially long on Sundays and short on Fridays (26.9 vs. 25.5 messages); 
more conversations are closed from midnight to 6 AM than from noon to 6 PM (75\% vs. 68\% of conversations). }

\subsubsection{Tractability}
Thus far, we have suggested that the counseling platform is representative of the inference task and challenges we formally described.
We additionally assume that the platform exhibits the properties under which these inference challenges can be mitigated; 
our framework and the validity of these assumptions were informed by interactions with the platform's staff and engineers. 
Nonetheless, 
\rrcomment{the assumptions we make are necessarily simplifying,
especially
as we do not directly control how the platform operates.}
For the purposes of demonstration, we argue that these assumptions are well-founded,
and revisit potential limitations in the discussion (Section~\ref{sec:discussion}).

\xhdr{Observed selection variables} 
We assume that the assignment of conversations to counselors is random up to the \textit{shift times} that counselors sign up to take.
As such, these shift times correspond to the fully-observable selection variables \varselectfull with which the bias from assignment can be addressed (Section \ref{sec:formulation_observation}).
This assumption reflects the platform's actual assignment process:
while counselors can choose which shifts to sign up for,
the platform assigns counselors to conversations randomly. 
For our demonstration we model shifts as temporal bins spanning the same 3-month window, day of week, and 6 hours of the day 
(e.g., Wednesdays from 12 to 6 AM, in January to March 2017).

\xhdr{Separable sets of conversations}
We also assume that there are no dependencies between different conversations taken by a counselor, allowing us to address the bias from interaction (Section \ref{sec:formulation_interaction}).
In particular, given the platform's focus on providing support in acute crises, texters generally do not contact the service repeatedly;
further, the platform does not deliberately assign repeat texters to the same counselor 
(i.e., in contrast to a therapy-oriented service).

\subsection{Analysis: Relating tendencies and outcomes}
\label{sec:demonstration_analysis}

\begin{figure*}[t]
\centering
\includegraphics[width=.98\textwidth]{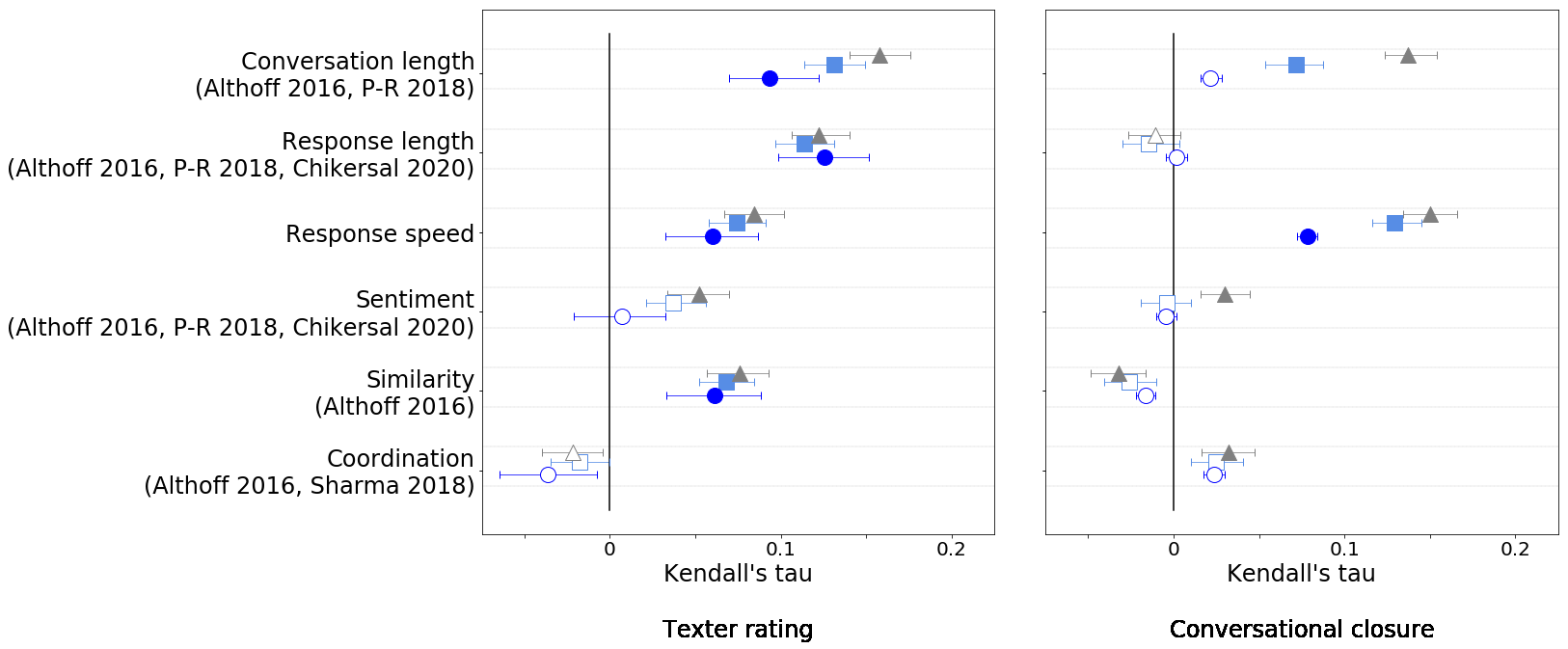}
\caption{\rredit{Relation between counselor-level behavioral tendencies and outcomes, measured as 
\KT
correlations, in increasingly controlled settings: \CorrelationalIcon  correlates counselor behavior and outcome propensity; \CrossIcon computes this correlation across temporally-interleaved splits of conversations; \PairedIcon further controls for shift time, 
thus reflecting the allocation effect formulated in Equation \ref{eq:basic_diff} while accounting for the inference challenges described in Sections \ref{sec:formulation_observation} and \ref{sec:formulation_interaction}. 
Error-bars show bootstrapped 95\% confidence intervals; shapes are filled for bootstrapped and Bonferroni-corrected $p < 0.01$. 
Abbreviated citations indicate studies that have demonstrated correlational relationships between the respective behaviors and outcomes.
}
}
\Description{Relation between counselor-level behavioral tendencies and outcomes, measured in increasingly controlled settings.}
\label{fig:kendall_tau}
\end{figure*}

In what follows, we use the counseling setting to empirically illustrate the inference challenges we formulated, as well as the solutions we proposed to address them.
As a demonstration, we examine the extent to which a representative selection of conversational tendencies
are related to the outcomes we've described,
 in terms of their allocation effect. 
\rredit{We consider three approaches to estimating the strength of the relation between a conversational tendency and outcome.
The estimates returned by each approach are denoted by different marker shapes in Figure \ref{fig:kendall_tau} (\CorrelationalIcon, \CrossIcon, \PairedIcon);
stronger effects are indicated by points further from the vertical line (indicating no effect).
The approaches are successively more rigorous in addressing the inference challenges, and contrasts in the effect sizes they estimate are visually represented as the different horizontal positions of the markers.
In comparing between these estimators,}
we show how a more careful analysis informed by our causal arguments 
\rredit{(represented as \PairedIcon)}
can distinguish between tendencies that could usefully guide an allocation policy, versus those that are related to outcome by virtue of circumstantial confounds
\rredit{(represented as \CorrelationalIcon and \CrossIcon)}, and that the platform may have less leverage over.

\xhdr{Conversational behaviors}
Past work in counseling has suggested a range of conversational
\rredit{behaviors which relate to counseling effectiveness
\cite{weger_active_2010,pyszczynski_depression_1987,mishara_which_2007,hill_clientcentered_2000,rogers_necessary_1957,rollnick_what_1995,haberstroh_experience_2007,richards_client-identified_2012,norcross_psychotherapy_2018}}.
Extending these efforts, recent computational studies have highlighted conversational features that 
\rredit{could signal positive outcomes in counseling conversations \cite{althoff_large-scale_2016,perez-rosas_understanding_2017,perez-rosas_analyzing_2018,cao_observing_2019,zhang_balancing_2020}, 
as well as other settings in the mental health domain like online support forums \cite{sharma_mental_2018,choudhury_language_2017} or longer-term therapy \cite{schroeder_data-driven_2020,chikersal_understanding_2020,pennebaker_psychological_2003}.}
The question of causality has 
\rrcomment{largely} been outside the scope of such studies;\footnote{\rredit{In support forum settings, 
\citet{choudhury_language_2017} and \citet{saha_causal_2020} similarly examine the causal effects of linguistic behaviors on outcomes such as the risk of suicidal ideation; the techniques they employ can be seen as controling for circumstance using observable attributes, which we contrast with our use of observed selection variables in Section \ref{sec:formulation_observation}.}}
here, we take up this question in terms of the allocation policy.
\rredit{For the purposes of demonstration, we focus on a small set
of behaviors, which we selected as simple representatives of prevalent types of conversational behaviors considered in these past works.
These behaviors are listed in Figure \ref{fig:kendall_tau}, along with studies 
which have demonstrated their correlations with mental health-related outcomes.
In particular, conversation length, response length and response speed relate to the \textit{fluency and pace} of the conversation \cite{althoff_large-scale_2016,perez-rosas_analyzing_2018,chikersal_understanding_2020};
sentiment is a frequently-cited attribute of the \textit{style or tone} of an utterance 
\cite[inter alia]{althoff_large-scale_2016,perez-rosas_analyzing_2018,chikersal_understanding_2020};
lexical similarity between utterances and linguistic coordination are often used to characterize \textit{interactional} behaviors 
\cite{sharma_mental_2018,althoff_large-scale_2016} like adapting to a client's language or reflecting their concerns.\footnote{\rredit{
We measure a counselor's speed in a conversation as the number of words they write, per minute taken to reply to a texter. Following \citet{althoff_large-scale_2016}, we measure sentiment as the VADER compound score of each message \cite{hutto_vader:_2014} and similarity as the cosine similarity between a counselor's message and the texter's preceding message; we obtain conversation-level measures of response length, sentiment and similarity by averaging over the counselor's messages in a conversation. 
As in \citet{althoff_large-scale_2016}, we use the approach in \cite{danescu-niculescu-mizil_echoes_2012} to measure coordination, noting that it produces a counselor-level, as opposed to a conversation-level score.}}}

Each of the behaviors we consider is potentially subject to the assignment- and interaction-based challenges we have described.
As shown in Figure \ref{fig:shift_time_variation}, they may be biased by a circumstantial factor like shift time.
\rredit{They may also reflect both the counselor's own conversational aptitude---e.g., their ability to manage conversational progress, maintain a helpful tone, and meaningfully respond to the texter---and the texter's inclinations---e.g., their responsiveness, emotional state, and openness to disclosing information.}
The subsequent analyses therefore clarify the extent to which the relations between these behaviors and outcomes, as surfaced by prior work, have causal interpretations in terms of the allocation effect.

\xhdr{Naive formulation: conversation-level effects}
\rrstat{We first compare counselor behaviors in conversations that are rated positively versus in those rated negatively, as well as in conversations that are closed versus in those where the texter disengages. 
For each conversation-level behavior and outcome, these comparisons yield statistically significant differences (Mann-Whitney U test $p < 0.01$), echoing several correlations reported in prior work between behaviors and outcomes in individual conversations.}
As we have argued, the usefulness of these relationships in guiding policies is unclear, since they could reflect circumstantial factors that the platform cannot influence.
\rrstat{For instance, the sentiment of counselor messages is significantly more positive in positively- versus negatively-rated conversations;
this could reflect the benefits of an upbeat tone, or that distressed texters who are harder to help 
also
tend to discuss less positive material.}
At the extreme, closed conversations are much longer than disengaged ones
\rrstat{(28.4 vs. 20.0 messages per conversation on average)},
perhaps tautologically: disengaged conversations end prematurely by definition.

\xhdr{Counselor-level correlations (\CorrelationalIcon)}
To build up to a counselor-level approach that addresses the influence of circumstance, \rrcomment{we first consider correlations between counselor-level aggregates of behavior \aggemptend{},\footnote{With the exception of coordination, which is already a counselor-level property, we derive counselor-level aggregates by averaging a counselor's per-conversation behaviors, e.g., average sentiment.} and of outcome \empout{} (computed as a counselor's proportion of positively-rated or closed conversations).}
\rrcomment{This view corresponds to the counselor-level approach taken in \citet{althoff_large-scale_2016}.\footnote{\rrcomment{Note that \citet{althoff_large-scale_2016} only consider the top and bottom 40 counselors in terms of \empout{}, while we consider all counselors.}}}

\rredit{To quantify the extent to which an aggregated behavior \aggemptend{} relates to an outcome propensity \empout{}, we compute 
\KT
correlations between \aggemptend{} and \empout{},
depicted in Figure \ref{fig:kendall_tau} as \CorrelationalIcon. 
At a high level, \KT compares the rankings of counselors according to \aggemptend{} and according to \empout{} by capturing the extent to which, within each pair of counselors, differences in \aggemptend{}  are in the same direction as differences in \empout{}. 
This mirrors our formulation of the allocation effect from Equation \ref{eq:basic_diff}, which is likewise defined over pairs of counselors;
here, however, we make the naive assumption that \aggemptend{} and \empout{} correspond to estimates of counselor tendency and outcome that can be meaningfully related 
(i.e., we ignore the two sources of bias).}

\xhdr{Addressing bias from interaction (\CrossIcon)} 
As described in Section \ref{sec:formulation_interaction}, both \aggemptend{} and \empout{} are entangled with the circumstances of conversations by virtue of the interaction between counselors and texters.
To mitigate the bias from interaction,
we divide each counselors' conversations into two splits, 
such that split \convosplit{0} consists of their even-indexed conversations (i.e., the second, fourth, sixth, \ldots) 
and split \convosplit{1} consists of their odd-indexed conversations.
\rrcomment{Using \KT,}
we compare the ranking of counselors according to their average behavior \aggemptend{\convosplit{0}} over split \convosplit{0} with their ranking based on their outcome propensity \empout{\convosplit{1}} over split \convosplit{1}, depicted as \CrossIcon in Figure \ref{fig:kendall_tau}.
\rrcomment{As such, \aggemptend{\convosplit{0}} and \empout{\convosplit{1}} correspond to estimates of counselor tendency and outcome which address this source of bias.}

\xhdr{Addressing bias from assignment (\PairedIcon)} 
The relation between 
\rrcomment{\aggemptend{\convosplit{0}} and \empout{\convosplit{1}} is still}
subject to biases incurred by the assignment of counselors and texters.
As discussed in Section \ref{sec:formulation_observation},
we address this problem by controling on shift time, our observed selection variable.
For each counselor \inter{J} and shift $s$, we compute a shift-specific outcome propensity \splitshiftempout{\inter{J}}{1}{s}
\rrcomment{(again over split \convosplit{1})}. 
For counselors \inter{J} and \inter{K}, we then compute the difference in outcome propensity over each shift they coincide on, 
$\shiftempeffectS{\obstend{\inter{J}}}{\obstend{\inter{K}}}{s} \!=\!\splitshiftempout{\inter{J}}{1}{s}\!-\!\splitshiftempout{\inter{K}}{1}{s}$. 
To aggregate across shifts, we take \empeffectS{\obstend{\inter{J}}}{\obstend{\inter{K}}} as the average of \shiftempeffectS{\obstend{\inter{J}}}{\obstend{\inter{K}}}{s} weighted by the number of conversations taken by the least-active of the two counselors within each shift.
Finally, we compute 
\KT
between outcome differences from split \convosplit{1} and behavioral patterns  from split \convosplit{0}, shown in Figure \ref{fig:kendall_tau} as \PairedIcon.\xspace
\footnote{\rredit{Our findings are qualitatively similar if we enforce that for each pair of counselors \inter{J} and \inter{K} considered in our measurement of the shift-controlled \KT statistic, the number of conversations they each take during shifts they are both in exceeds some minimum number; for the rating outcome, this minimum pertains to the number of rated conversations they each take in such shifts.
Choosing smaller thresholds potentially incurs noisier measurements of outcome propensity,
while more restrictive cut-offs result in less statistical power; 
in the case of rating, our analyses would be limited to a potentially skewed sample of shifts and of counselors where enough ratings are obtained.}} 

Having addressed the two sources of bias, the values represented as \PairedIcon thus 
\rrcomment{correspond to an unbiased estimate of}
the allocation effect of the conversational tendencies we've examined.

\xhdr{Results} 
\rredit{In comparing different counselor-level approaches, we highlight the two sources of bias that are in play, and show how addressing them can moderate our understanding of how different tendencies and outcomes are related.}
\rredit{This is depicted in Figure \ref{fig:kendall_tau} as \PairedIcons which are hollow---indicating no statistical significance---or closer to the vertical line (at \KT$=0$) than corresponding \CorrelationalIcons or \CrossIcons---indicating that the latter two approaches overestimated the effect size.}
For example, the large counselor-level effects of conversation length on closure (\CorrelationalIcon) diminish drastically  after addressing interactional bias (\CrossIcon), showing that length tautologically reflects closure. 
\rrcomment{Further addressing the temporally-mediated assignment bias (\PairedIcon) shows that this tendency does not have a significant effect on closure;
the decreased effect size also suggests that the previously-observed relation may also have been contingent on shift time, echoing the across-shift variations in conversation length depicted in Figure \ref{fig:shift_time_variation}.}

\rrcomment{These results distinguish between tendencies that are potentially useful in guiding allocation policies, and those that are correlated to outcome by way of circumstantial factors.
While these correlations suggest that such tendencies could be highly informative of a conversation's circumstances, they do not translate to recommendations for how the platform should allocate counselors.
For example, while conversations in which a counselor's utterances exhibited more positive sentiment resulted in better outcomes, this is largely due to circumstantial and interactional effects,
perhaps reflecting texters who are easier to help, a priori any interaction. 
As such, we should not expect that allocating counselors with a tendency to use more positive language to a conversation will increase the likelihood of receiving a positive rating or closing properly.
In contrast, allocating more conversations to counselors who tend to write longer messages or to better echo the texters (higher similarity) may be more promising in improving ratings.}

\subsection{\rredit{Simulated experiment: Estimating the effects of an allocation policy}}
\label{sec:demonstration_simulation}

\rredit{Having demonstrated how our framework can be used to probe the causal nature of the relation between behavioral tendencies and conversational outcomes, we now estimate the potential impact of an allocation policy that assigns counselors to conversations based on these tendencies.
We do this by \emph{simulating} an implementation of the allocation policy under idealized conditions, following prior work in the medical domain \cite{kleinberg_prediction_2015}.
We note that this simulated experiment is only meant to serve as a feasibility check that could precede 
(but not stand in for)
more costly real-world experimentation 
on the actual platform (e.g., via randomized controlled trials \cite{angrist_mostly_2008,rubin_design_2007}). 
}

\rredit{In practice, to implement and test a tendency-based allocation policy, the platform could proceed in two steps.
First, it would identify counselors who exhibit conversational tendencies with a positive allocation effect on an outcome, using our framework to ensure that measurements of these effects aren't biased by the inference challenges we've described.
We simulate this step by translating the empirical approach  detailed in the previous section to train a \textit{predictive model} that can then be used to identify counselors who are likely to be effective in future conversations (\S \ref{sec:simulation_prediction}).}

\rredit{Second the platform would allocate counselors to incoming texters based on their predicted effectiveness, comparing the resultant outcomes to a control condition (e.g., outcomes in a random subset of shifts where counselors are still allocated randomly).
We coarsely simulate this step by estimating the effects of a counterfactual re-allocation of counselors within each shift, based on their predicted efficiency (\S \ref{sec:simulation_reallocation}).}

\subsubsection{Training a predictive model}
\label{sec:simulation_prediction}
\rrcomment{We first identify potentially effective counselors in terms of the rating and closure outcomes. Here, we translate the empirical approach from Section \ref{sec:demonstration_analysis} into the setup of a predictive task; 
for the purposes of demonstration, we use the simple conversational tendencies from the preceding analysis as features, 
\rredit{noting that future work could naturally consider richer representations of tendencies}.}

\xhdr{Task setup}
\rrcomment{Crucially, in modeling counselor effectiveness, we must ensure that the assignment and interaction-based inference challenges are both addressed. 
To do so for multiple tendencies at once, we integrate our inference solutions into the setup of a prediction task:}
given a counselor's \rrcomment{behavioral tendencies exhibited} in their first 40 conversations---i.e., their \textit{past behavior}---we wish to predict their propensity to receive good ratings or close conversations in their next 40 conversations---i.e., their \textit{future outcomes}.
In other words, we \rrcomment{\emph{split}} counselors' conversations into past and future conversations, relating tendencies estimated from one split to outcome propensities estimated from the other and thus mitigating bias from interaction.

\rrcomment{To additionally address bias from assignment we formulate this prediction task as a \emph{paired} task that matches counselors on our selection variable, shift. 
Concretely, for each pair of counselors that have future conversations in a given shift, our goal is to guess which counselor will get a higher proportion of good outcomes in that shift. 
}

We \rrcomment{divide} the population of counselors into a train and test set comprising \rrcomment{50\% of counselors each}.
\rrcomment{We train the model for texter rating over 10,118 pairs spanning 280 shifts; for closure we train on 55,473 pairs across 329 shifts.}
We use SVM models with 10-fold cross-validation.

\xhdr{Model performance}
Before applying these models to \rredit{our simulated experiment,} we report their performance \rrcomment{for reference.}
The relative test set accuracy of the trained models corroborates the effect sizes we observed following our controlled approach (denoted by \PairedIcon in Figure \ref{fig:kendall_tau}).
In predicting \emph{future} rating, features based on the minimalistic set of tendencies we used outperform a random (50\%) baseline (Bonferroni-corrected binomial test $p \!<\! 0.001$) with an accuracy of \rrcomment{56.6\%}.
In predicting future closure rate, these features get a lower accuracy of \rrcomment{50.8\%}---the poorer performance suggests that the apparently strong naive correlations (corresponding to \CorrelationalIcon in Figure \ref{fig:kendall_tau}, right) are contingent on the circumstantial confounds that our task setup addresses.\xspace
\footnote{\rrcomment{The relative magnitude of the feature weights learned by each model echo the ranking of the controled assignment effects depicted as \PairedIcon in Figure \ref{fig:kendall_tau}. 
For instance, in the closure task, conversation length has the smallest weight while response speed has the largest (in spite of similar counselor-level correlations denoted by \CorrelationalIcon).}}
Therefore, we expect behavioral tendencies to be less informative 
\rrcomment{in improving the closure outcome via an assignment policy,
than in improving the rating outcome.}

\subsubsection{\rredit{Simulated re-allocation}}

\label{sec:simulation_reallocation}

\rredit{We now use observational data to \textit{simulate} an allocation policy that assigns counselors that are predicted to be more effective---based on their behavioral tendencies exhibited in \emph{past} conversations---to more \emph{future} conversations in their shifts.
Estimating the impact of this policy requires addressing the same counterfactual question introduced in Section \ref{sec:formulation}---given counselors with different tendencies, what is the effect of allocating one counselor to a conversation, versus the other?
Here, we adapt a procedure from prior work that simulates the effectiveness of 
policies in a medical domain \cite{kleinberg_prediction_2015}.}

\xhdr{\rredit{Simplifying assumptions}} 
\rredit{Before detailing our simulated policy, we note that it draws on the key assumption that within-shift re-allocations are feasible.
In practice, the logistical and ethical aspects of this assumption would need to be carefully considered: such a policy would hinge on the availability and willingness of the counselors, their capacity for taking more conversations, and the potentially detrimental burden they would incur from the extra load.\footnote{We do not consider cross-shift reallocations since those would potentially imply requiring counselors to work at times at which they are not available.}
For the purposes of this demonstration, this assumption allows us to focus 
on gauging the allocation effect that we have explored in the preceding sections; as such, we highlight the causal inference problem while leaving important and complementary aspects of the policy to future work.
}

\xhdr{\rredit{Counterfactual re-allocation}} 
Given a shift, we focus on the subset of conversations taken by the test-set counselors
during their 40th to 80th (i.e., future) conversations.
\rrcomment{We use our paired prediction models to produce rankings of these counselors based on tendencies observed in their first 40 (i.e., past) conversations,
and in separate shifts from their 40th to 80th conversations. 
}
We consider a counterfactual scenario in which all conversations in this shift are taken by the top $k\%$ of counselors according to these predicted rankings,
\rrcomment{thus simulating allocating more conversations to them and making use of our assumption that counselors can be re-allocated within a shift.}

\begin{figure}[t]
\centering
\includegraphics[width=.98\textwidth]{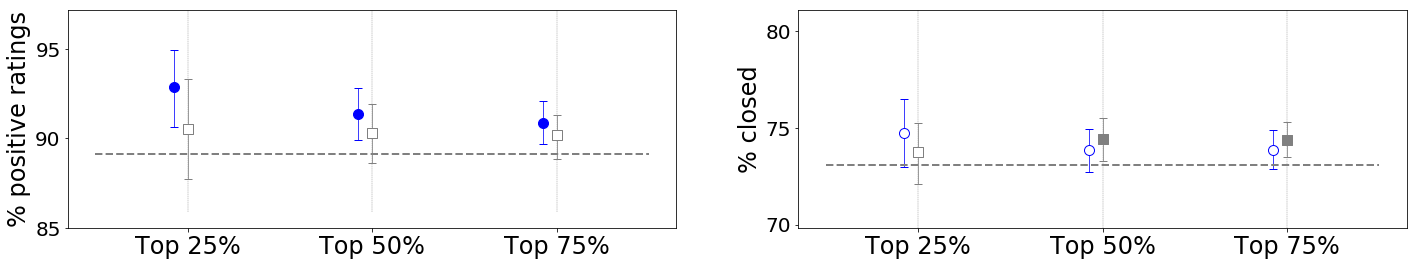}
\caption{\rredit{Proportions of positive ratings (left) or closed conversations (right), macroaveraged over shifts, in a simulated counterfactual setting where the platform selects the top $k\%$ counselors on the basis of their future performance as predicted using past conversational tendencies \textcolor{blue}{$\bigcirc$}, or based on historical outcomes \textcolor{gray}{$\square$}. 
Error bars show bootstrapped 95\% confidence intervals.
The dashed line denotes the actual realized proportion of positive ratings or proper closures observed in the data.
Shapes are filled in for Bonferroni-corrected Wilcoxon $p < 0.01$, comparing counterfactual and realized outcomes and pairing on shift.}}
\Description{Proportions of positive ratings or closed conversations in simulated counterfactual settings.}
\label{fig:replacement_task}
\end{figure}

To estimate the effect of this within-shift re-allocation, we compare the proportion of good outcomes over conversations taken by these predicted top counselors---the \textit{counterfactual outcome}---to the actual proportion---the \textit{realized outcome}.
\rrcomment{
We macro-average these within-shift outcomes across all shifts, so that no single shift has a disproportionate impact on the estimate.
}
\rrcomment{To ensure that we have enough data to provide clear estimates, the statistics we subsequently report are taken over shifts with at least six counselors who each take at least three of their 40th to 80th conversations in that shift; 
for the rating outcome we further enforce that each of these conversations receives a rating from the texter.
As such, we consider 55 shifts for rating and 138 for closure.\footnote{\rrcomment{Small modifications to these parameters yield qualitatively similar results. In choosing these parameters, we acknowledge some tradeoffs:
lowering these thresholds incurs some noise---concretely, the counterfactual performances would be taken over a fewer number of counselors and conversations, increasing the chance that the estimates are skewed by exceptionally good or bad predictions or outcomes. Raising these thresholds decreases the number of shifts considered, resulting in a loss of statistical power and potentially constraining the representativeness of the findings.}}}
\xhdr{Estimated effects}
Figure~\ref{fig:replacement_task} shows 
counterfactual outcomes, macroaveraged by shift, for different $k$ (\textcolor{blue}{$\bigcirc$}), compared to the realized outcome (dashed line).
For texter rating, each counterfactual outcome 
\rrcomment{improves upon a realized outcome.}
\rredit{Pairing on shift, these differences are significant for each $k$ (Wilcoxon $p < 0.01$, indicated as filled-in \textcolor{blue}{$\CIRCLE$}), suggesting that the counterfactual improvements occur across many different shifts:
concretely, at $k=25\%$, the counterfactual outcome improves over the realized one in 74\% of shifts.}
This suggests that there is some promise in allocation policies that are informed by conversational tendencies,
\rredit{and could motivate more involved studies, such as those deploying experiments that more actively intervene in the platform.}

For the closure outcome, the counterfactual scenario does not improve significantly over the realized one 
\rrcomment{(Wilcoxon $p \geq 0.01$ for each, indicated as \textcolor{blue}{$\bigcirc$}).}
\rredit{This further reinforces that the strong relationships between tendency and closure reflected in naive approaches (\CorrelationalIcon and \CrossIcon in Figure \ref{fig:kendall_tau}), which do not have a causal interpretation, cannot usefully guide allocation policies.
As such, more involved experimentation may be unwarranted.}

\subsubsection{Comparison to non-conversational information}

\rrcomment{As an alternative to gauging the effectiveness of conversational tendencies, the platform may wish to rely on other information that might more directly relate to outcomes.
\rredit{In particular, a counselor's \textbf{past outcomes}---i.e., their propensity to elicit positive ratings or to close conversations, as computed over their first 40 conversations---could be a strong signal of their future effectiveness.}
Here, we briefly evaluate the utility of these non-conversational signals in guiding the re-allocation.}

\xhdr{\rrcomment{Results}}
\rrcomment{Following the same simulation procedure as before, we start by identifying potentially effective counselors on the basis of past outcomes.
For rating, a predictive model based on past ratings gets an accuracy of 53.1\%---while this outperforms the random baseline, it underperforms relative to the model trained on tendency (Binomial $p<0.001$ for both). 
\rredit{This suggests that past outcomes are indicative of future performance (having accounted for our inference challenges in the prediction task setup). }
Nonetheless, the conversational tendencies we considered---while simplistic---are more informative than non-conversational indicators; 
the counselors' behavioral tendencies may be less noisy and more robust across circumstances than the past ratings they receive from different texters, especially given the low response rate.}

\rrcomment{Reflecting the relatively low prediction accuracies, the counterfactual outcomes from accounting for past ratings do not significantly improve upon realized outcomes (Wilcoxon $p \geq 0.01$, results macroaveraged over shift shown in Figure 6 left, as \textcolor{gray}{$\square$}). 
At each $k$, these outcomes are lower than the counterfactual outcomes attained using the tendency-based approach (comparing relative heights of the \textcolor{blue}{$\CIRCLE$} and \textcolor{gray}{$\square$} markers), 
\rredit{though none of these differences are statistically significant (Wilcoxon $p \geq 0.01$).
This shows that there is some promise to going beyond non-conversational information, while motivating the need for richer conversational signals that could be more informative.}}

\rrcomment{For closure,
a predictive model based on a counselor's past ability to close conversations slightly outperforms the model using tendencies, with an accuracy of 51.6\% ($p < 0.001$). 
We find that at $k=50\%$ and $75\%$, the counterfactual outcomes from using past closure propensities do significantly exceed realized outcomes (Wilcoxon $p < 0.01$,  Figure 6 right, indicated as \textcolor{gray}{$\blacksquare$}).\footnote{We attribute the lower counterfactual estimate at $k=25\%$ to noise, since fewer counselors are re-allocated at this setting.} 
}

\section{Discussion}
\label{sec:discussion}
In this work, we considered the problem of translating observed relationships between conversational behaviors and outcomes into actionable insights.
Through examining a particular policy---allocating agents in conversational platforms---that such observational analyses could inform,
we formally described the inference task and inherent challenges involved, 
translating causal inference arguments to the domain of conversations.
In the context of crisis counseling, we  demonstrated the importance of properly addressing these challenges, but also the potential benefits of policies that are informed by careful analyses of conversations.
Here, we describe how our particular work informs broader studies of computer-mediated settings where conversations play a central role.
We also highlight some empirical and theoretical limitations that future work could take up. 
\rredit{The bulk of our work examines one type of conversational setting---goal-oriented asymmetric conversation platforms---and one type of policy---allocating agents. 
This focus enables us to develop theoretical descriptions that clearly highlight key aspects of the inference task---the relationships between behavior, circumstance, and outcome;
it is also grounded in a socially consequential real-world setting, crisis counseling.
We note, however, that this inference task is relevant to a broader range of conversational settings: 
like team discussions or public forms, 
where conversational roles may be more fluid and extend beyond that of agent and client,
while different participants may have different goals in the conversation.
A platform could also pursue a range of other policies, such as \textit{training} effective conversational behaviors 
or deterring detrimental ones via practices like moderation.
These related settings and policies inherit the challenges we have described: the underlying problems of causally relating behaviors and outcomes, and of addressing the inference challenges from assignment and interaction, continue to be salient beyond our particular focus.}

We see our work as critically examining one important part of developing policies to improve conversational platforms: 
translating descriptive findings based on observational data into prescriptive information.
Designing and implementing these policies requires a wealth of additional research.
For instance, other studies could examine more intricate conversational behaviors and tendencies, 
viewing the rich types of characteristics explored in many past descriptive studies through a causal lens.
A more nuanced understanding
of conversational outcomes is needed to gauge the effectiveness of any policy.
\rredit{For example, in this setting, a complementary line of work could develop more informative ways to elicit texter feedback, seeking to better understand and mitigate the presently low response rate; understanding longer-term impacts of a counseling conversation would also be valuable.}

Numerous factors relating to the implementation of a policy would need to be accounted for as well;
\rredit{as noted earlier, the estimated benefits of allocating more conversations to counselors must be 
weighed against the potential for the additional workload to strain the counselors' mental health and conversational effectiveness.}
Addressing these aspects is beyond the scope of our work, and 
we look to other studies of computer-mediated communication platforms for promising approaches \cite{kraut_building_2012,kittur_future_2013}.
However, we emphasize that the core problem of measuring the causal effects of a policy is salient regardless of the extent to which other components of the policy are well-developed---\rredit{such that identifying these effects, while not sufficient, is necessary in informing these policies.}

\subsection{Limitations}

Our present study is subject to two broad types of limitations, 
relating to the extent to which our conceptual description is a good model of real-world communication platforms like the counseling service,
and to aspects of the model that could be extended to encompass a broader range of settings.

\xhdr{Empirical limitations}
Throughout our paper, we've argued that the type of conversation platform we theoretically examined, and the assumptions necessary to mitigate inference challenges, 
are represented in the crisis counseling setting
and are realistic across a broader range of domains.
As we've noted, we  must inevitably make some assumptions about the nature of the counseling platform.
Concretely, we supposed for sake of demonstration that assignment is random within shift and that dependencies do not exist between different conversations taken by a counselor. 
In practice, the platform's assignment procedure prioritizes clear cases of suicidal ideation, and more experienced counselors can take multiple conversations at once, which may induce cross-conversational dependencies.
Future work could better account for factors such as these, that exist across other conversational settings.
In particular, these efforts could investigate the extent to which the solutions we propose are sensitive to these exceptions,
and could relax the assumptions that these exceptions challenge.

\xhdr{Theoretical limitations}
Our theoretical formulation---of a particular inference problem in a particular type of conversational setting---could be relaxed or extended in several ways.
In the purview of this subproblem, the solution we propose to mitigate the bias from assignment requires us to control for a selector variable such as shift, and hence restricts us from making comparisons or agent allocations across shifts,
\rrcomment{or from applying our approach to settings where selector variables are not fully observable.}
Future work could consider ways to relax this requirement, perhaps by using parametric models of the relation between shift, behavior and outcome.

We have examined a simple formulation of an allocation policy: discovering and hence allocating more conversations to the most effective agents within a shift. 
Future work could examine more sophisticated allocation policies, and the more complex causal inferences required to motivate them.
For instance, finer-grained models of tendency and circumstance could point to policies that match agents with situations they are particularly well-suited to address, given their behavioral tendencies.
\rredit{A complementary line of work could more rigorously interrogate the assumption at the heart of our simulated re-allocation (\S \ref{sec:simulation_reallocation}) by accounting for the impact 
of increasing agents' loads,
 or by examining policies that are less contingent on this assumption.}

Our methodology could also be extended to encompass a broader range of conversational paradigms beyond asymmetric settings,
such those involving discussions in a team \cite{hansen_benefits_2006,cheng_what_2019,whiting_did_2019,baer_personality_2008}.
\rredit{Such extensions would need to account for additional properties of the domain, such as a more diverse and dynamic range of conversational roles, over which the platform might have varying degrees of leverage.}
In such contexts, the attributes of the individuals involved as gleaned from indicators such as personality types or demographic information, as well as how these attributes are combined, has been experimentally shown to potentially impact the effectiveness of a discussion \cite{bear_role_2011,woolley_what_2011}.
Here, it would be interesting to see how conversational tendencies could be used in concert with these non-interactional labels.

Our work is crucially limited to addressing the problem of agent-level allocation.
As such, we've examined a coarser policy than that of training agents to adopt particular behaviors. 
Future work could take up the corresponding inference task: estimating how a change in an agent's behavior, once they are assigned to a conversation, affects the conversation's outcome.
Intuitively, this task is more challenging to address: 
as previously noted, taking agent-level aggregates allows us to abstract away from the circumstances within a conversation in our analyses;
in the training paradigm, circumstantial factors could be even more intricately entangled with behavior and outcome.

Our work addresses two key features of analyzing conversational data:
that this data is often observational,
and that it contains complex interactional dynamics.
However, we leave an additional pillar of this setting open: conversations are \textit{linguistic}.
As such, the behavioral signals we glean from the data are necessarily low-dimensional representations \cite{egami_how_2018,wang_when_2019,keith_text_2020} of abstract and perhaps more consequential conversational qualities.
For instance, our formulation allows us to reason about the effect of allocating agents, and in the counseling setting we show that verbosity is a good \textit{signal} of a counselor's effectiveness.
However, unilaterally instructing to counselors that they increase the number of words they use may be ineffectual, if verbosity is simply a proxy for a tendency that is less straightforward to model, such as eloquence.
\rredit{This present limitation further constrains our ability to translate inferences we've made to policies such as training, and to 
make finer-grained statements
about conversational behaviors and their impacts.}
Thus, there is ample opportunity for future work to address this gap, by way of more nuanced examinations of conversational behaviors and of their causal effects.

\begin{acks}

The authors would like to thank the participants at the New Directions in Analyzing Text as Data conference (Fall 2019) and the Interdisciplinary Seminar in Quantitative Methods at the University of Michigan (Fall 2019), as well as Jonathan P. Chang, Caleb Chiam, Liye Fu and Lillian Lee for helpful discussions. This research would not have been possible without the support of Crisis Text Line, and we are particularly grateful to Robert Filbin, Christine Morrison, and Jaclyn Weiser for their valuable insights into the platform and for their help with using the data. The research has been supported in part by NSF CAREER Award IIS1750615 and a Microsoft Research PhD Fellowship. The collaboration with Crisis Text Line was supported by the Robert Wood Johnson Foundation; the views expressed here do not necessarily reflect the views of the foundation.

\end{acks}

\bibliographystyle{ACM-Reference-Format}
\bibliography{ctl-tendencies-tacl-autoupdate-JZ}


\begin{thebibliography}{72}


\ifx \showCODEN    \undefined \def \showCODEN     #1{\unskip}     \fi
\ifx \showDOI      \undefined \def \showDOI       #1{#1}\fi
\ifx \showISBNx    \undefined \def \showISBNx     #1{\unskip}     \fi
\ifx \showISBNxiii \undefined \def \showISBNxiii  #1{\unskip}     \fi
\ifx \showISSN     \undefined \def \showISSN      #1{\unskip}     \fi
\ifx \showLCCN     \undefined \def \showLCCN      #1{\unskip}     \fi
\ifx \shownote     \undefined \def \shownote      #1{#1}          \fi
\ifx \showarticletitle \undefined \def \showarticletitle #1{#1}   \fi
\ifx \showURL      \undefined \def \showURL       {\relax}        \fi
\providecommand\bibfield[2]{#2}
\providecommand\bibinfo[2]{#2}
\providecommand\natexlab[1]{#1}
\providecommand\showeprint[2][]{arXiv:#2}

\bibitem[\protect\citeauthoryear{Althoff, Clark, and Leskovec}{Althoff
  et~al\mbox{.}}{2016}]%
        {althoff_large-scale_2016}
\bibfield{author}{\bibinfo{person}{Tim Althoff}, \bibinfo{person}{Kevin Clark},
  {and} \bibinfo{person}{Jure Leskovec}.} \bibinfo{year}{2016}\natexlab{}.
\newblock \showarticletitle{Large-Scale {{Analysis}} of {{Counseling
  Conversations}}: {{An Application}} of {{Natural Language Processing}} to
  {{Mental Health}}}.
\newblock \bibinfo{journal}{\emph{Transactions of the Association for
  Computational Linguistics}} (\bibinfo{year}{2016}).
\newblock


\bibitem[\protect\citeauthoryear{Andrade, Bagri, Zaw, Roos, and Ruiz}{Andrade
  et~al\mbox{.}}{2010}]%
        {andrade_avatar-mediated_2010}
\bibfield{author}{\bibinfo{person}{Allen~D. Andrade}, \bibinfo{person}{Anita
  Bagri}, \bibinfo{person}{Khin Zaw}, \bibinfo{person}{Bernard~A. Roos}, {and}
  \bibinfo{person}{Jorge~G. Ruiz}.} \bibinfo{year}{2010}\natexlab{}.
\newblock \showarticletitle{Avatar-Mediated Training in the Delivery of Bad
  News in a Virtual World}.
\newblock \bibinfo{journal}{\emph{Journal of Palliative Medicine}}
  (\bibinfo{year}{2010}).
\newblock


\bibitem[\protect\citeauthoryear{Angrist and Pischke}{Angrist and
  Pischke}{2008}]%
        {angrist_mostly_2008}
\bibfield{author}{\bibinfo{person}{Joshua~D. Angrist} {and}
  \bibinfo{person}{J{\"o}rn-Steffen Pischke}.} \bibinfo{year}{2008}\natexlab{}.
\newblock \bibinfo{booktitle}{\emph{Mostly {{Harmless Econometrics}}: {{An
  Empiricist}}'s {{Companion}}}}.
\newblock \bibinfo{publisher}{{Princeton University Press}}.
\newblock


\bibitem[\protect\citeauthoryear{Angrist and Pischke}{Angrist and
  Pischke}{2010}]%
        {angrist_credibility_2010}
\bibfield{author}{\bibinfo{person}{Joshua~D. Angrist} {and}
  \bibinfo{person}{J{\"o}rn-Steffen Pischke}.} \bibinfo{year}{2010}\natexlab{}.
\newblock \showarticletitle{The {{Credibility Revolution}} in {{Empirical
  Economics}}: {{How Better Research Design Is Taking}} the {{Con}} out of
  {{Econometrics}}}.
\newblock \bibinfo{journal}{\emph{Journal of Economic Perspectives}}
  (\bibinfo{year}{2010}).
\newblock


\bibitem[\protect\citeauthoryear{Arguello, Butler, Joyce, Kraut, Ling,
  Ros{\'e}, and Wang}{Arguello et~al\mbox{.}}{2006}]%
        {arguello_talk_2006}
\bibfield{author}{\bibinfo{person}{Jaime Arguello}, \bibinfo{person}{Brian~S.
  Butler}, \bibinfo{person}{Elisabeth Joyce}, \bibinfo{person}{Robert Kraut},
  \bibinfo{person}{Kimberly~S. Ling}, \bibinfo{person}{Carolyn Ros{\'e}}, {and}
  \bibinfo{person}{Xiaoqing Wang}.} \bibinfo{year}{2006}\natexlab{}.
\newblock \showarticletitle{Talk to Me: Foundations for Successful
  Individual-Group Interactions in Online Communities}. In
  \bibinfo{booktitle}{\emph{Proceedings of {{CHI}}}}.
\newblock


\bibitem[\protect\citeauthoryear{Baer, Oldham, Jacobsohn, and
  Hollingshead}{Baer et~al\mbox{.}}{2008}]%
        {baer_personality_2008}
\bibfield{author}{\bibinfo{person}{Markus Baer}, \bibinfo{person}{Greg~R.
  Oldham}, \bibinfo{person}{Gwendolyn~Costa Jacobsohn}, {and}
  \bibinfo{person}{Andrea~B. Hollingshead}.} \bibinfo{year}{2008}\natexlab{}.
\newblock \showarticletitle{The {{Personality Composition}} of {{Teams}} and
  {{Creativity}}: {{The Moderating Role}} of {{Team Creative Confidence}}}.
\newblock \bibinfo{journal}{\emph{The Journal of Creative Behavior}}
  (\bibinfo{year}{2008}).
\newblock


\bibitem[\protect\citeauthoryear{Bear and Woolley}{Bear and Woolley}{2011}]%
        {bear_role_2011}
\bibfield{author}{\bibinfo{person}{Julia Bear} {and} \bibinfo{person}{Anita
  Woolley}.} \bibinfo{year}{2011}\natexlab{}.
\newblock \showarticletitle{The {{Role}} of {{Gender}} in {{Team
  Collaboration}} and {{Performance}}}.
\newblock \bibinfo{journal}{\emph{Interdisciplinary Science Reviews}}
  (\bibinfo{year}{2011}).
\newblock


\bibitem[\protect\citeauthoryear{Brito and Pearl}{Brito and Pearl}{2012}]%
        {brito_generalized_2012}
\bibfield{author}{\bibinfo{person}{Carlos Brito} {and} \bibinfo{person}{Judea
  Pearl}.} \bibinfo{year}{2012}\natexlab{}.
\newblock \showarticletitle{Generalized {{Instrumental Variables}}}. In
  \bibinfo{booktitle}{\emph{Proceedings of {{UAI}}}}.
\newblock


\bibitem[\protect\citeauthoryear{Burke, Joyce, Kim, Anand, and Kraut}{Burke
  et~al\mbox{.}}{2007}]%
        {burke_introductions_2007}
\bibfield{author}{\bibinfo{person}{Moira Burke}, \bibinfo{person}{Elisabeth
  Joyce}, \bibinfo{person}{Tackjin Kim}, \bibinfo{person}{Vivek Anand}, {and}
  \bibinfo{person}{Robert Kraut}.} \bibinfo{year}{2007}\natexlab{}.
\newblock \showarticletitle{Introductions and {{Requests}}: {{Rhetorical
  Strategies That Elicit Response}} in {{Online Communities}}}. In
  \bibinfo{booktitle}{\emph{Communities and {{Technologies}} 2007}},
  \bibfield{editor}{\bibinfo{person}{Charles Steinfield},
  \bibinfo{person}{Brian~T. Pentland}, \bibinfo{person}{Mark Ackerman}, {and}
  \bibinfo{person}{Noshir Contractor}} (Eds.). \bibinfo{publisher}{{Springer}}.
\newblock


\bibitem[\protect\citeauthoryear{Cao, Tanana, Imel, Poitras, Atkins, and
  Srikumar}{Cao et~al\mbox{.}}{2019}]%
        {cao_observing_2019}
\bibfield{author}{\bibinfo{person}{Jie Cao}, \bibinfo{person}{Michael Tanana},
  \bibinfo{person}{Zac Imel}, \bibinfo{person}{Eric Poitras},
  \bibinfo{person}{David Atkins}, {and} \bibinfo{person}{Vivek Srikumar}.}
  \bibinfo{year}{2019}\natexlab{}.
\newblock \showarticletitle{Observing {{Dialogue}} in {{Therapy}}:
  {{Categorizing}} and {{Forecasting Behavioral Codes}}}. In
  \bibinfo{booktitle}{\emph{Proceedings of {{ACL}}}}.
\newblock


\bibitem[\protect\citeauthoryear{Chancellor, Hu, and De~Choudhury}{Chancellor
  et~al\mbox{.}}{2018}]%
        {chancellor_norms_2018}
\bibfield{author}{\bibinfo{person}{Stevie Chancellor}, \bibinfo{person}{Andrea
  Hu}, {and} \bibinfo{person}{Munmun De~Choudhury}.}
  \bibinfo{year}{2018}\natexlab{}.
\newblock \showarticletitle{Norms {{Matter}}: {{Contrasting Social Support
  Around Behavior Change}} in {{Online Weight Loss Communities}}}. In
  \bibinfo{booktitle}{\emph{Proceedings of {{CHI}}}}.
\newblock


\bibitem[\protect\citeauthoryear{Cheng, Yang, Tan, Cheng, Zhuang, and
  Cheng}{Cheng et~al\mbox{.}}{2019}]%
        {cheng_what_2019}
\bibfield{author}{\bibinfo{person}{Ziqiang Cheng}, \bibinfo{person}{Yang Yang},
  \bibinfo{person}{Chenhao Tan}, \bibinfo{person}{Denny Cheng},
  \bibinfo{person}{Yueting Zhuang}, {and} \bibinfo{person}{Alex Cheng}.}
  \bibinfo{year}{2019}\natexlab{}.
\newblock \showarticletitle{What {{Makes}} a {{Good Team}}? {{A Large}}-Scale
  {{Study}} on the {{Effect}} of {{Team Composition}} in {{Honor}} of
  {{Kings}}}. In \bibinfo{booktitle}{\emph{Proceedings of {{WWW}}}}.
\newblock


\bibitem[\protect\citeauthoryear{Chikersal, Belgrave, Doherty, Enrique,
  Palacios, Richards, and Thieme}{Chikersal et~al\mbox{.}}{2020}]%
        {chikersal_understanding_2020}
\bibfield{author}{\bibinfo{person}{Prerna Chikersal}, \bibinfo{person}{Danielle
  Belgrave}, \bibinfo{person}{Gavin Doherty}, \bibinfo{person}{Angel Enrique},
  \bibinfo{person}{Jorge~E. Palacios}, \bibinfo{person}{Derek Richards}, {and}
  \bibinfo{person}{Anja Thieme}.} \bibinfo{year}{2020}\natexlab{}.
\newblock \showarticletitle{Understanding {{Client Support Strategies}} to
  {{Improve Clinical Outcomes}} in an {{Online Mental Health Intervention}}}.
  In \bibinfo{booktitle}{\emph{Proceedings of {{CHI}}}}.
\newblock


\bibitem[\protect\citeauthoryear{Choudhury and K{\i}c{\i}man}{Choudhury and
  K{\i}c{\i}man}{2017}]%
        {choudhury_language_2017}
\bibfield{author}{\bibinfo{person}{Munmun~De Choudhury} {and}
  \bibinfo{person}{Emre K{\i}c{\i}man}.} \bibinfo{year}{2017}\natexlab{}.
\newblock \showarticletitle{The {{Language}} of {{Social Support}} in {{Social
  Media}} and Its {{Effect}} on {{Suicidal Ideation Risk}}}. In
  \bibinfo{booktitle}{\emph{Proceedings of {{ICWSM}}}}.
\newblock


\bibitem[\protect\citeauthoryear{Cranshaw and Kittur}{Cranshaw and
  Kittur}{2011}]%
        {cranshaw_polymath_2011}
\bibfield{author}{\bibinfo{person}{Justin Cranshaw} {and}
  \bibinfo{person}{Aniket Kittur}.} \bibinfo{year}{2011}\natexlab{}.
\newblock \showarticletitle{The Polymath Project: Lessons from a Successful
  Online Collaboration in Mathematics}. In
  \bibinfo{booktitle}{\emph{Proceedings of {{CHI}}}}.
\newblock


\bibitem[\protect\citeauthoryear{{Danescu-Niculescu-Mizil}, Lee, Pang, and
  Kleinberg}{{Danescu-Niculescu-Mizil} et~al\mbox{.}}{2012}]%
        {danescu-niculescu-mizil_echoes_2012}
\bibfield{author}{\bibinfo{person}{Cristian {Danescu-Niculescu-Mizil}},
  \bibinfo{person}{Lillian Lee}, \bibinfo{person}{Bo Pang}, {and}
  \bibinfo{person}{Jon Kleinberg}.} \bibinfo{year}{2012}\natexlab{}.
\newblock \showarticletitle{Echoes of {{Power}}: {{Language Effects}} and
  {{Power Differences}} in {{Social Interaction}}}. In
  \bibinfo{booktitle}{\emph{Proceedings of {{WWW}}}}.
\newblock


\bibitem[\protect\citeauthoryear{DeMasi, Hearst, and Recht}{DeMasi
  et~al\mbox{.}}{2019}]%
        {demasi_towards_2019}
\bibfield{author}{\bibinfo{person}{Orianna DeMasi}, \bibinfo{person}{Marti~A.
  Hearst}, {and} \bibinfo{person}{Benjamin Recht}.}
  \bibinfo{year}{2019}\natexlab{}.
\newblock \showarticletitle{Towards {{Augmenting Crisis Counselor Training}} by
  {{Improving Message Retrieval}}}.
\newblock \bibinfo{journal}{\emph{Proceedings of the Workshop on Computational
  Linguistics and Clinical Psychology}} (\bibinfo{year}{2019}).
\newblock


\bibitem[\protect\citeauthoryear{Egami, Fong, Grimmer, Roberts, and
  Stewart}{Egami et~al\mbox{.}}{2018}]%
        {egami_how_2018}
\bibfield{author}{\bibinfo{person}{Naoki Egami}, \bibinfo{person}{Christian~J.
  Fong}, \bibinfo{person}{Justin Grimmer}, \bibinfo{person}{Margaret~E.
  Roberts}, {and} \bibinfo{person}{Brandon~M. Stewart}.}
  \bibinfo{year}{2018}\natexlab{}.
\newblock \bibinfo{title}{How to {{Make Causal Inferences Using Texts}}}.
  (\bibinfo{year}{2018}).
\newblock


\bibitem[\protect\citeauthoryear{Fountain}{Fountain}{2003}]%
        {fountain_prospects_2003}
\bibfield{author}{\bibinfo{person}{Jane~E. Fountain}.}
  \bibinfo{year}{2003}\natexlab{}.
\newblock \showarticletitle{Prospects for Improving the Regulatory Process
  Using E-Rulemaking}.
\newblock \bibinfo{journal}{\emph{Commun. ACM}} (\bibinfo{year}{2003}).
\newblock


\bibitem[\protect\citeauthoryear{Graesser, Person, and Magliano}{Graesser
  et~al\mbox{.}}{1995}]%
        {graesser_collaborative_1995}
\bibfield{author}{\bibinfo{person}{Arthur~C. Graesser},
  \bibinfo{person}{Natalie~K. Person}, {and} \bibinfo{person}{Joseph~P.
  Magliano}.} \bibinfo{year}{1995}\natexlab{}.
\newblock \showarticletitle{Collaborative Dialogue Patterns in Naturalistic
  One-to-One Tutoring}.
\newblock \bibinfo{journal}{\emph{Applied Cognitive Psychology}}
  (\bibinfo{year}{1995}).
\newblock


\bibitem[\protect\citeauthoryear{Haberstroh, Duffey, Evans, Gee, and
  Trepal}{Haberstroh et~al\mbox{.}}{2007}]%
        {haberstroh_experience_2007}
\bibfield{author}{\bibinfo{person}{Shane Haberstroh}, \bibinfo{person}{Thelma
  Duffey}, \bibinfo{person}{Marcheta Evans}, \bibinfo{person}{Robert Gee},
  {and} \bibinfo{person}{Heather Trepal}.} \bibinfo{year}{2007}\natexlab{}.
\newblock \showarticletitle{The {{Experience}} of {{Online Counseling}}}.
\newblock \bibinfo{journal}{\emph{Journal of Mental Health Counseling}}
  (\bibinfo{year}{2007}).
\newblock


\bibitem[\protect\citeauthoryear{Hansen}{Hansen}{2006}]%
        {hansen_benefits_2006}
\bibfield{author}{\bibinfo{person}{Randall~S. Hansen}.}
  \bibinfo{year}{2006}\natexlab{}.
\newblock \showarticletitle{Benefits and {{Problems}} with {{Student Teams}}:
  {{Suggestions}} for {{Improving Team Projects}}.}
\newblock \bibinfo{journal}{\emph{Journal of Education for Business}}
  (\bibinfo{year}{2006}).
\newblock


\bibitem[\protect\citeauthoryear{Hill and Nakayama}{Hill and Nakayama}{2000}]%
        {hill_clientcentered_2000}
\bibfield{author}{\bibinfo{person}{Clara~E. Hill} {and}
  \bibinfo{person}{Emilie~Y. Nakayama}.} \bibinfo{year}{2000}\natexlab{}.
\newblock \showarticletitle{Client-centered Therapy: {{Where}} Has It Been and
  Where Is It Going? {{A}} Comment on {{Hathaway}} (1948)}.
\newblock \bibinfo{journal}{\emph{Journal of Clinical Psychology}}
  (\bibinfo{year}{2000}).
\newblock


\bibitem[\protect\citeauthoryear{Hu, Tafti, and Gal}{Hu et~al\mbox{.}}{2019}]%
        {hu_read_2019}
\bibfield{author}{\bibinfo{person}{Yuheng Hu}, \bibinfo{person}{Ali Tafti},
  {and} \bibinfo{person}{David Gal}.} \bibinfo{year}{2019}\natexlab{}.
\newblock \showarticletitle{Read {{This}}, {{Please}}? {{The Role}} of
  {{Politeness}} in {{Customer Service Engagement}} on {{Social Media}}}. In
  \bibinfo{booktitle}{\emph{Proceedings of {{HICSS}}}}.
\newblock


\bibitem[\protect\citeauthoryear{Hutto and Gilbert}{Hutto and Gilbert}{2014}]%
        {hutto_vader:_2014}
\bibfield{author}{\bibinfo{person}{Clayton~J. Hutto} {and}
  \bibinfo{person}{Eric Gilbert}.} \bibinfo{year}{2014}\natexlab{}.
\newblock \showarticletitle{{{VADER}}: {{A Parsimonious Rule}}-Based {{Model}}
  for {{Sentiment Analysis}} of {{Social Media Text}}}. In
  \bibinfo{booktitle}{\emph{Proceedings of {{ICWSM}}}}.
\newblock


\bibitem[\protect\citeauthoryear{Im, Zhang, Schilling, and Karger}{Im
  et~al\mbox{.}}{2018}]%
        {im_deliberation_2018}
\bibfield{author}{\bibinfo{person}{Jane Im}, \bibinfo{person}{Amy~X. Zhang},
  \bibinfo{person}{Christopher~J. Schilling}, {and} \bibinfo{person}{David
  Karger}.} \bibinfo{year}{2018}\natexlab{}.
\newblock \showarticletitle{Deliberation and {{Resolution}} on {{Wikipedia}}:
  {{A Case Study}} of {{Requests}} for {{Comments}}}. In
  \bibinfo{booktitle}{\emph{Proceedings of {{CSCW}}}}.
\newblock


\bibitem[\protect\citeauthoryear{Jaech, Zayats, Fang, Ostendorf, and
  Hajishirzi}{Jaech et~al\mbox{.}}{2015}]%
        {jaech_talking_2015}
\bibfield{author}{\bibinfo{person}{Aaron Jaech}, \bibinfo{person}{Victoria
  Zayats}, \bibinfo{person}{Hao Fang}, \bibinfo{person}{Mari Ostendorf}, {and}
  \bibinfo{person}{Hannaneh Hajishirzi}.} \bibinfo{year}{2015}\natexlab{}.
\newblock \showarticletitle{Talking to the Crowd: {{What}} Do People React to
  in Online Discussions?}. In \bibinfo{booktitle}{\emph{Proceedings of
  {{EMNLP}}}}.
\newblock


\bibitem[\protect\citeauthoryear{Keith, Jensen, and O'Connor}{Keith
  et~al\mbox{.}}{2020}]%
        {keith_text_2020}
\bibfield{author}{\bibinfo{person}{Katherine~A. Keith}, \bibinfo{person}{David
  Jensen}, {and} \bibinfo{person}{Brendan O'Connor}.}
  \bibinfo{year}{2020}\natexlab{}.
\newblock \showarticletitle{Text and {{Causal Inference}}: {{A Review}} of
  {{Using Text}} to {{Remove Confounding}} from {{Causal Estimates}}}. In
  \bibinfo{booktitle}{\emph{Proceedings of {{ACL}}}}.
\newblock


\bibitem[\protect\citeauthoryear{Kim, Engel, Woolley, Lin, McArthur, and
  Malone}{Kim et~al\mbox{.}}{2017}]%
        {kim_what_2017}
\bibfield{author}{\bibinfo{person}{Young~Ji Kim}, \bibinfo{person}{David
  Engel}, \bibinfo{person}{Anita~Williams Woolley}, \bibinfo{person}{Jeffrey
  Yu-Ting Lin}, \bibinfo{person}{Naomi McArthur}, {and}
  \bibinfo{person}{Thomas~W. Malone}.} \bibinfo{year}{2017}\natexlab{}.
\newblock \showarticletitle{What {{Makes}} a {{Strong Team}}?: {{Using
  Collective Intelligence}} to {{Predict Team Performance}} in {{League}} of
  {{Legends}}}. In \bibinfo{booktitle}{\emph{Proceedings of {{CSCW}}}}.
\newblock


\bibitem[\protect\citeauthoryear{Kittur, Nickerson, Bernstein, Gerber, Shaw,
  Zimmerman, Lease, and Horton}{Kittur et~al\mbox{.}}{2013}]%
        {kittur_future_2013}
\bibfield{author}{\bibinfo{person}{Aniket Kittur}, \bibinfo{person}{Jeffrey~V.
  Nickerson}, \bibinfo{person}{Michael Bernstein}, \bibinfo{person}{Elizabeth
  Gerber}, \bibinfo{person}{Aaron Shaw}, \bibinfo{person}{John Zimmerman},
  \bibinfo{person}{Matt Lease}, {and} \bibinfo{person}{John Horton}.}
  \bibinfo{year}{2013}\natexlab{}.
\newblock \showarticletitle{The Future of Crowd Work}. In
  \bibinfo{booktitle}{\emph{Proceedings of {{CSCW}}}}.
\newblock


\bibitem[\protect\citeauthoryear{Kleinberg, Ludwig, Mullainathan, and
  Obermeyer}{Kleinberg et~al\mbox{.}}{2015}]%
        {kleinberg_prediction_2015}
\bibfield{author}{\bibinfo{person}{Jon Kleinberg}, \bibinfo{person}{Jens
  Ludwig}, \bibinfo{person}{Sendhil Mullainathan}, {and} \bibinfo{person}{Ziad
  Obermeyer}.} \bibinfo{year}{2015}\natexlab{}.
\newblock \showarticletitle{Prediction {{Policy Problems}}}.
\newblock \bibinfo{journal}{\emph{American Economic Review}}
  (\bibinfo{year}{2015}).
\newblock


\bibitem[\protect\citeauthoryear{Kraut and Resnick}{Kraut and Resnick}{2012}]%
        {kraut_building_2012}
\bibfield{author}{\bibinfo{person}{Robert~E. Kraut} {and} \bibinfo{person}{Paul
  Resnick}.} \bibinfo{year}{2012}\natexlab{}.
\newblock \bibinfo{booktitle}{\emph{Building {{Successful Online Communities}}:
  {{Evidence}}-{{Based Social Design}}}}.
\newblock \bibinfo{publisher}{{MIT Press}}.
\newblock


\bibitem[\protect\citeauthoryear{Lasecki, White, Murray, and Bigham}{Lasecki
  et~al\mbox{.}}{2012}]%
        {lasecki_crowd_2012}
\bibfield{author}{\bibinfo{person}{Walter~S. Lasecki},
  \bibinfo{person}{Samuel~C. White}, \bibinfo{person}{Kyle~I. Murray}, {and}
  \bibinfo{person}{Jeffrey~P. Bigham}.} \bibinfo{year}{2012}\natexlab{}.
\newblock \showarticletitle{Crowd {{Memory}}: {{Learning}} in the
  {{Collective}}}. In \bibinfo{booktitle}{\emph{Proceedings of {{CHI}}}}.
\newblock


\bibitem[\protect\citeauthoryear{Luther, Caine, Ziegler, and Bruckman}{Luther
  et~al\mbox{.}}{2010}]%
        {luther_why_2010}
\bibfield{author}{\bibinfo{person}{Kurt Luther}, \bibinfo{person}{Kelly Caine},
  \bibinfo{person}{Kevin Ziegler}, {and} \bibinfo{person}{Amy Bruckman}.}
  \bibinfo{year}{2010}\natexlab{}.
\newblock \showarticletitle{Why It Works (When It Works): {{Success}} Factors
  in Online Creative Collaboration}. In \bibinfo{booktitle}{\emph{Proceedings
  of {{GROUP}}}}.
\newblock


\bibitem[\protect\citeauthoryear{Lykourentzou, Antoniou, Naudet, and
  Dow}{Lykourentzou et~al\mbox{.}}{2016}]%
        {lykourentzou_personality_2016-1}
\bibfield{author}{\bibinfo{person}{I. Lykourentzou}, \bibinfo{person}{Angeliki
  Antoniou}, \bibinfo{person}{Yannick Naudet}, {and} \bibinfo{person}{Steven
  Dow}.} \bibinfo{year}{2016}\natexlab{}.
\newblock \showarticletitle{Personality {{Matters}}: {{Balancing}} for
  {{Personality Types Leads}} to {{Better Outcomes}} for {{Crowd Teams}}}. In
  \bibinfo{booktitle}{\emph{Proceedings of {{CSCW}}}}.
\newblock


\bibitem[\protect\citeauthoryear{Maki, Yoder, Jo, and Ros{\'e}}{Maki
  et~al\mbox{.}}{2017}]%
        {maki_roles_2017}
\bibfield{author}{\bibinfo{person}{Keith Maki}, \bibinfo{person}{Michael
  Yoder}, \bibinfo{person}{Yohan Jo}, {and} \bibinfo{person}{Carolyn
  Ros{\'e}}.} \bibinfo{year}{2017}\natexlab{}.
\newblock \showarticletitle{Roles and {{Success}} in {{Wikipedia Talk Pages}}:
  {{Identifying Latent Patterns}} of {{Behavior}}}. In
  \bibinfo{booktitle}{\emph{Proceedings of {{EMNLP}}}}.
\newblock


\bibitem[\protect\citeauthoryear{McInnis, Cosley, Baumer, and Leshed}{McInnis
  et~al\mbox{.}}{2018}]%
        {mcinnis_effects_2018}
\bibfield{author}{\bibinfo{person}{Brian McInnis}, \bibinfo{person}{Dan
  Cosley}, \bibinfo{person}{Eric Baumer}, {and} \bibinfo{person}{Gilly
  Leshed}.} \bibinfo{year}{2018}\natexlab{}.
\newblock \showarticletitle{Effects of {{Comment Curation}} and {{Opposition}}
  on {{Coherence}} in {{Online Policy Discussion}}}. In
  \bibinfo{booktitle}{\emph{Proceedings of {{GROUP}}}}.
\newblock


\bibitem[\protect\citeauthoryear{Mishara, Chagnon, Daigle, Balan, Raymond,
  Marcoux, Bardon, Campbell, and Berman}{Mishara et~al\mbox{.}}{2007}]%
        {mishara_which_2007}
\bibfield{author}{\bibinfo{person}{Brian~L. Mishara}, \bibinfo{person}{Fran{\c
  c}ois Chagnon}, \bibinfo{person}{Marc~S. Daigle}, \bibinfo{person}{Bogdan
  Balan}, \bibinfo{person}{Sylvaine Raymond}, \bibinfo{person}{Isabelle
  Marcoux}, \bibinfo{person}{C{\'e}cile Bardon}, \bibinfo{person}{Julie~K.
  Campbell}, {and} \bibinfo{person}{Alan~D. Berman}.}
  \bibinfo{year}{2007}\natexlab{}.
\newblock \showarticletitle{Which Helper Behaviors and Intervention Styles Are
  Related to Better Short-Term Outcomes in Telephone Crisis Intervention?
  {{Results}} from a {{Silent Monitoring Study}} of {{Calls}} to the
  {{U}}.{{S}}. 1-800-{{SUICIDE Network}}.}
\newblock \bibinfo{journal}{\emph{Suicide \& Life-Threatening Behavior}}
  (\bibinfo{year}{2007}).
\newblock


\bibitem[\protect\citeauthoryear{Niculae and {Danescu-Niculescu-Mizil}}{Niculae
  and {Danescu-Niculescu-Mizil}}{2016}]%
        {niculae_conversational_2016}
\bibfield{author}{\bibinfo{person}{Vlad Niculae} {and}
  \bibinfo{person}{Cristian {Danescu-Niculescu-Mizil}}.}
  \bibinfo{year}{2016}\natexlab{}.
\newblock \showarticletitle{Conversational {{Markers}} of {{Constructive
  Discussions}}}. In \bibinfo{booktitle}{\emph{Proceedings of {{NAACL}}}}.
\newblock


\bibitem[\protect\citeauthoryear{Norcross and Lambert}{Norcross and
  Lambert}{2018}]%
        {norcross_psychotherapy_2018}
\bibfield{author}{\bibinfo{person}{John~C. Norcross} {and}
  \bibinfo{person}{Michael~J. Lambert}.} \bibinfo{year}{2018}\natexlab{}.
\newblock \showarticletitle{Psychotherapy Relationships That Work {{III}}.}
\newblock \bibinfo{journal}{\emph{Psychotherapy}} (\bibinfo{year}{2018}).
\newblock


\bibitem[\protect\citeauthoryear{Packard, Moore, and McFerran}{Packard
  et~al\mbox{.}}{2018}]%
        {packard_im_2018}
\bibfield{author}{\bibinfo{person}{Grant Packard}, \bibinfo{person}{Sarah~G.
  Moore}, {and} \bibinfo{person}{Brent McFerran}.}
  \bibinfo{year}{2018}\natexlab{}.
\newblock \showarticletitle{({{I}}'m) {{Happy}} to {{Help}} ({{You}}): {{The
  Impact}} of {{Personal Pronoun Use}} in {{Customer}}\textendash{{Firm
  Interactions}}}.
\newblock \bibinfo{journal}{\emph{Journal of Marketing Research}}
  (\bibinfo{year}{2018}).
\newblock


\bibitem[\protect\citeauthoryear{Pavalanathan and Eisenstein}{Pavalanathan and
  Eisenstein}{2015}]%
        {pavalanathan_emoticons_2015}
\bibfield{author}{\bibinfo{person}{Umashanthi Pavalanathan} {and}
  \bibinfo{person}{Jacob Eisenstein}.} \bibinfo{year}{2015}\natexlab{}.
\newblock \showarticletitle{Emoticons vs. {{Emojis}} on {{Twitter}}: {{A Causal
  Inference Approach}}}. In \bibinfo{booktitle}{\emph{Proceedings of
  {{OSSM}}}}.
\newblock


\bibitem[\protect\citeauthoryear{Pearl}{Pearl}{1995}]%
        {pearl_causal_1995}
\bibfield{author}{\bibinfo{person}{Judea Pearl}.}
  \bibinfo{year}{1995}\natexlab{}.
\newblock \showarticletitle{Causal Diagrams for Empirical Research}.
\newblock \bibinfo{journal}{\emph{Biometrika}} (\bibinfo{year}{1995}).
\newblock


\bibitem[\protect\citeauthoryear{Pearl}{Pearl}{2013}]%
        {pearl_testability_2013}
\bibfield{author}{\bibinfo{person}{Judea Pearl}.}
  \bibinfo{year}{2013}\natexlab{}.
\newblock \showarticletitle{On the {{Testability}} of {{Causal Models}} with
  {{Latent}} and {{Instrumental Variables}}}. In
  \bibinfo{booktitle}{\emph{Proceedings of {{UAI}}}}.
\newblock


\bibitem[\protect\citeauthoryear{Pennebaker, Mehl, and Niederhoffer}{Pennebaker
  et~al\mbox{.}}{2003}]%
        {pennebaker_psychological_2003}
\bibfield{author}{\bibinfo{person}{James~W. Pennebaker},
  \bibinfo{person}{Matthias~R. Mehl}, {and} \bibinfo{person}{Kate~G.
  Niederhoffer}.} \bibinfo{year}{2003}\natexlab{}.
\newblock \showarticletitle{Psychological {{Aspects}} of {{Natural Language
  Use}}: {{Our Words}}, {{Our Selves}}}.
\newblock \bibinfo{journal}{\emph{Annual Review of Psychology}}
  (\bibinfo{year}{2003}).
\newblock


\bibitem[\protect\citeauthoryear{{P{\'e}rez-Rosas}, Mihalcea, Resnicow, Singh,
  and An}{{P{\'e}rez-Rosas} et~al\mbox{.}}{2017}]%
        {perez-rosas_understanding_2017}
\bibfield{author}{\bibinfo{person}{Ver{\'o}nica {P{\'e}rez-Rosas}},
  \bibinfo{person}{Rada Mihalcea}, \bibinfo{person}{Kenneth Resnicow},
  \bibinfo{person}{Satinder Singh}, {and} \bibinfo{person}{Lawrence An}.}
  \bibinfo{year}{2017}\natexlab{}.
\newblock \showarticletitle{Understanding and {{Predicting Empathic Behavior}}
  in {{Counseling Therapy}}}. In \bibinfo{booktitle}{\emph{Proceedings of
  {{ACL}}}}.
\newblock


\bibitem[\protect\citeauthoryear{{P{\'e}rez-Rosas}, Sun, Li, Wang, Resnicow,
  and Mihalcea}{{P{\'e}rez-Rosas} et~al\mbox{.}}{2018}]%
        {perez-rosas_analyzing_2018}
\bibfield{author}{\bibinfo{person}{Ver{\'o}nica {P{\'e}rez-Rosas}},
  \bibinfo{person}{Xuetong Sun}, \bibinfo{person}{Christy Li},
  \bibinfo{person}{Yuchen Wang}, \bibinfo{person}{Kenneth Resnicow}, {and}
  \bibinfo{person}{Rada Mihalcea}.} \bibinfo{year}{2018}\natexlab{}.
\newblock \showarticletitle{Analyzing the {{Quality}} of {{Counseling
  Conversations}}: The {{Tell}}-{{Tale Signs}} of {{High}}-Quality
  {{Counseling}}}. In \bibinfo{booktitle}{\emph{Proceedings of {{LREC}}}}.
\newblock


\bibitem[\protect\citeauthoryear{Pisani, Kanuri, Filbin, Gallo, Gould, Lehmann,
  Levine, Marcotte, Pascal, Rousseau, Turner, Yen, and Ranney}{Pisani
  et~al\mbox{.}}{2019}]%
        {pisani_protecting_2019}
\bibfield{author}{\bibinfo{person}{Anthony~R. Pisani}, \bibinfo{person}{Nitya
  Kanuri}, \bibinfo{person}{Bob Filbin}, \bibinfo{person}{Carlos Gallo},
  \bibinfo{person}{Madelyn Gould}, \bibinfo{person}{Lisa~S. Lehmann},
  \bibinfo{person}{Robert Levine}, \bibinfo{person}{John~E. Marcotte},
  \bibinfo{person}{Brian Pascal}, \bibinfo{person}{David Rousseau},
  \bibinfo{person}{Shairi Turner}, \bibinfo{person}{Shirley Yen}, {and}
  \bibinfo{person}{Megan~L. Ranney}.} \bibinfo{year}{2019}\natexlab{}.
\newblock \showarticletitle{Protecting {{User Privacy}} and {{Rights}} in
  {{Academic Data}}-{{Sharing Partnerships}}: {{Principles From}} a {{Pilot
  Program}} at {{Crisis Text Line}}}.
\newblock \bibinfo{journal}{\emph{Journal of Medical Internet Research}}
  (\bibinfo{year}{2019}).
\newblock


\bibitem[\protect\citeauthoryear{Pyszczynski, Holt, and Greenberg}{Pyszczynski
  et~al\mbox{.}}{1987}]%
        {pyszczynski_depression_1987}
\bibfield{author}{\bibinfo{person}{Tom Pyszczynski}, \bibinfo{person}{Kathleen
  Holt}, {and} \bibinfo{person}{Jeff Greenberg}.}
  \bibinfo{year}{1987}\natexlab{}.
\newblock \showarticletitle{Depression, Self-Focused Attention, and
  Expectancies for Positive and Negative Future Life Events for Self and
  Others}.
\newblock \bibinfo{journal}{\emph{Journal of Personality and Social
  Psychology}} (\bibinfo{year}{1987}).
\newblock


\bibitem[\protect\citeauthoryear{Richards and Timulak}{Richards and
  Timulak}{2012}]%
        {richards_client-identified_2012}
\bibfield{author}{\bibinfo{person}{Derek Richards} {and}
  \bibinfo{person}{Ladislav Timulak}.} \bibinfo{year}{2012}\natexlab{}.
\newblock \showarticletitle{Client-Identified Helpful and Hindering Events in
  Therapist-Delivered vs. Self-Administered Online Cognitive-Behavioural
  Treatments for Depression in College Students}.
\newblock \bibinfo{journal}{\emph{Counselling Psychology Quarterly}}
  (\bibinfo{year}{2012}).
\newblock


\bibitem[\protect\citeauthoryear{Rogers}{Rogers}{1957}]%
        {rogers_necessary_1957}
\bibfield{author}{\bibinfo{person}{Carl~R. Rogers}.}
  \bibinfo{year}{1957}\natexlab{}.
\newblock \showarticletitle{The Necessary and Sufficient Conditions of
  Therapeutic Personality Change}.
\newblock \bibinfo{journal}{\emph{Journal of Consulting Psychology}}
  (\bibinfo{year}{1957}).
\newblock


\bibitem[\protect\citeauthoryear{Rollnick and Miller}{Rollnick and
  Miller}{1995}]%
        {rollnick_what_1995}
\bibfield{author}{\bibinfo{person}{Stephen Rollnick} {and}
  \bibinfo{person}{William~R. Miller}.} \bibinfo{year}{1995}\natexlab{}.
\newblock \showarticletitle{What Is {{Motivational Interviewing}}?}
\newblock \bibinfo{journal}{\emph{Behavioural and Cognitive Psychotherapy}}
  (\bibinfo{year}{1995}).
\newblock


\bibitem[\protect\citeauthoryear{Rosenbaum}{Rosenbaum}{2010}]%
        {rosenbaum_design_2010}
\bibfield{author}{\bibinfo{person}{Paul~R. Rosenbaum}.}
  \bibinfo{year}{2010}\natexlab{}.
\newblock \bibinfo{booktitle}{\emph{Design of Observational Studies}}.
\newblock \bibinfo{publisher}{{Springer}}.
\newblock


\bibitem[\protect\citeauthoryear{Rubin}{Rubin}{2007}]%
        {rubin_design_2007}
\bibfield{author}{\bibinfo{person}{Donald~B. Rubin}.}
  \bibinfo{year}{2007}\natexlab{}.
\newblock \showarticletitle{The Design versus the Analysis of Observational
  Studies for Causal Effects: Parallels with the Design of Randomized Trials}.
\newblock \bibinfo{journal}{\emph{Stat. Med.}} (\bibinfo{year}{2007}).
\newblock


\bibitem[\protect\citeauthoryear{Saha and Sharma}{Saha and Sharma}{2020}]%
        {saha_causal_2020}
\bibfield{author}{\bibinfo{person}{Koustuv Saha} {and} \bibinfo{person}{Amit
  Sharma}.} \bibinfo{year}{2020}\natexlab{}.
\newblock \showarticletitle{Causal {{Factors}} of {{Effective Psychosocial
  Outcomes}} in {{Online Mental Health Communities}}}. In
  \bibinfo{booktitle}{\emph{Proceeding of {{ICWSM}}}}.
\newblock


\bibitem[\protect\citeauthoryear{Schroeder, Suh, Wilks, Czerwinski, Munson,
  Fogarty, and Althoff}{Schroeder et~al\mbox{.}}{2020}]%
        {schroeder_data-driven_2020}
\bibfield{author}{\bibinfo{person}{Jessica Schroeder}, \bibinfo{person}{Jina
  Suh}, \bibinfo{person}{Chelsey Wilks}, \bibinfo{person}{Mary Czerwinski},
  \bibinfo{person}{Sean~A. Munson}, \bibinfo{person}{James Fogarty}, {and}
  \bibinfo{person}{Tim Althoff}.} \bibinfo{year}{2020}\natexlab{}.
\newblock \showarticletitle{Data-{{Driven Implications}} for {{Translating
  Evidence}}-{{Based Psychotherapies}} into {{Technology}}-{{Delivered
  Interventions}}}. In \bibinfo{booktitle}{\emph{Proceedings of
  {{PervasiveHealth}}}}.
\newblock


\bibitem[\protect\citeauthoryear{Schwalbe, Oh, and Zweben}{Schwalbe
  et~al\mbox{.}}{2014}]%
        {schwalbe_sustaining_2014}
\bibfield{author}{\bibinfo{person}{Craig~S. Schwalbe}, \bibinfo{person}{Hans~Y.
  Oh}, {and} \bibinfo{person}{Allen Zweben}.} \bibinfo{year}{2014}\natexlab{}.
\newblock \showarticletitle{Sustaining Motivational Interviewing: A
  Meta-Analysis of Training Studies}.
\newblock \bibinfo{journal}{\emph{Addiction}} (\bibinfo{year}{2014}).
\newblock


\bibitem[\protect\citeauthoryear{Sharma and De~Choudhury}{Sharma and
  De~Choudhury}{2018}]%
        {sharma_mental_2018}
\bibfield{author}{\bibinfo{person}{Eva Sharma} {and} \bibinfo{person}{Munmun
  De~Choudhury}.} \bibinfo{year}{2018}\natexlab{}.
\newblock \showarticletitle{Mental {{Health Support}} and Its {{Relationship}}
  to {{Linguistic Accommodation}} in {{Online Communities}}}. In
  \bibinfo{booktitle}{\emph{Proceedings of {{CHI}}}}.
\newblock


\bibitem[\protect\citeauthoryear{Sridhar and Getoor}{Sridhar and
  Getoor}{2019}]%
        {sridhar_estimating_2019}
\bibfield{author}{\bibinfo{person}{Dhanya Sridhar} {and} \bibinfo{person}{Lise
  Getoor}.} \bibinfo{year}{2019}\natexlab{}.
\newblock \showarticletitle{Estimating {{Causal Effects}} of {{Tone}} in
  {{Online Debates}}}. In \bibinfo{booktitle}{\emph{Proceedings of {{IJCAI}}}}.
\newblock


\bibitem[\protect\citeauthoryear{Tan, Niculae, {Danescu-Niculescu}, and
  Lee}{Tan et~al\mbox{.}}{2016}]%
        {tan_winning_2016}
\bibfield{author}{\bibinfo{person}{Chenhao Tan}, \bibinfo{person}{Vlad
  Niculae}, \bibinfo{person}{Cristian {Danescu-Niculescu}}, {and}
  \bibinfo{person}{Lillian Lee}.} \bibinfo{year}{2016}\natexlab{}.
\newblock \showarticletitle{Winning {{Arguments}}: {{Interaction Dynamics}} and
  {{Persuasion Strategies}} in {{Good}}-Faith {{Online Discussions}}}. In
  \bibinfo{booktitle}{\emph{Proceedings of {{WWW}}}}.
\newblock


\bibitem[\protect\citeauthoryear{Tracey, Wampold, Lichtenberg, and
  Goodyear}{Tracey et~al\mbox{.}}{2014}]%
        {tracey_expertise_2014}
\bibfield{author}{\bibinfo{person}{Terence J.~G. Tracey},
  \bibinfo{person}{Bruce~E. Wampold}, \bibinfo{person}{James~W. Lichtenberg},
  {and} \bibinfo{person}{Rodney~K. Goodyear}.} \bibinfo{year}{2014}\natexlab{}.
\newblock \showarticletitle{Expertise in Psychotherapy: An Elusive Goal?}
\newblock \bibinfo{journal}{\emph{The American Psychologist}}
  (\bibinfo{year}{2014}).
\newblock


\bibitem[\protect\citeauthoryear{Wang, Kraut, and Levine}{Wang
  et~al\mbox{.}}{2012}]%
        {wang_stay_2012}
\bibfield{author}{\bibinfo{person}{Yi-chia Wang}, \bibinfo{person}{Robert
  Kraut}, {and} \bibinfo{person}{John~M. Levine}.}
  \bibinfo{year}{2012}\natexlab{}.
\newblock \showarticletitle{To {{Stay}} or {{Leave}}? {{The Relationship}} of
  {{Emotional}} and {{Informational Support}} to {{Commitment}} in {{Online
  Health Support Groups}}}. In \bibinfo{booktitle}{\emph{Proceeding of
  {{CSCW}}}}.
\newblock


\bibitem[\protect\citeauthoryear{Wang and Culotta}{Wang and Culotta}{2019}]%
        {wang_when_2019}
\bibfield{author}{\bibinfo{person}{Zhao Wang} {and} \bibinfo{person}{Aron
  Culotta}.} \bibinfo{year}{2019}\natexlab{}.
\newblock \showarticletitle{When Do {{Words Matter}}? {{Understanding}} the
  {{Impact}} of {{Lexical Choice}} on {{Audience Perception}} Using
  {{Individual Treatment Effect Estimation}}}. In
  \bibinfo{booktitle}{\emph{Proceedings of {{AAAI}}}}.
\newblock


\bibitem[\protect\citeauthoryear{Weger, Castle, and Emmett}{Weger
  et~al\mbox{.}}{2010}]%
        {weger_active_2010}
\bibfield{author}{\bibinfo{person}{Harry Weger}, \bibinfo{person}{Gina~R.
  Castle}, {and} \bibinfo{person}{Melissa~C. Emmett}.}
  \bibinfo{year}{2010}\natexlab{}.
\newblock \showarticletitle{Active {{Listening}} in {{Peer Interviews}}: {{The
  Influence}} of {{Message Paraphrasing}} on {{Perceptions}} of {{Listening
  Skill}}}.
\newblock \bibinfo{journal}{\emph{International Journal of Listening}}
  (\bibinfo{year}{2010}).
\newblock


\bibitem[\protect\citeauthoryear{Whiting, Blaising, Barreau, Fiuza, Marda,
  Valentine, and Bernstein}{Whiting et~al\mbox{.}}{2019}]%
        {whiting_did_2019}
\bibfield{author}{\bibinfo{person}{Mark~E. Whiting}, \bibinfo{person}{Allie
  Blaising}, \bibinfo{person}{Chloe Barreau}, \bibinfo{person}{Laura Fiuza},
  \bibinfo{person}{Nik Marda}, \bibinfo{person}{Melissa Valentine}, {and}
  \bibinfo{person}{Michael~S. Bernstein}.} \bibinfo{year}{2019}\natexlab{}.
\newblock \showarticletitle{Did {{It Have To End This Way}}?: {{Understanding
  The Consistency}} of {{Team Fracture}}}. In
  \bibinfo{booktitle}{\emph{Proceedings of {{CHI}}}}.
\newblock


\bibitem[\protect\citeauthoryear{Woolley and Malone}{Woolley and
  Malone}{2011}]%
        {woolley_what_2011}
\bibfield{author}{\bibinfo{person}{Anita Woolley} {and} \bibinfo{person}{Thomas
  Malone}.} \bibinfo{year}{2011}\natexlab{}.
\newblock \showarticletitle{What Makes a Team Smarter? {{More}} Women}.
\newblock \bibinfo{journal}{\emph{Harvard Business Review}}
  (\bibinfo{year}{2011}).
\newblock


\bibitem[\protect\citeauthoryear{Yang, Halfaker, Kraut, and Hovy}{Yang
  et~al\mbox{.}}{2016}]%
        {yang_who_2016}
\bibfield{author}{\bibinfo{person}{Diyi Yang}, \bibinfo{person}{Aaron
  Halfaker}, \bibinfo{person}{Robert Kraut}, {and} \bibinfo{person}{Eduard
  Hovy}.} \bibinfo{year}{2016}\natexlab{}.
\newblock \showarticletitle{Who {{Did What}}: {{Editor Role Identification}} in
  {{Wikipedia}}}. In \bibinfo{booktitle}{\emph{Proceedings of {{ICWSM}}}}.
\newblock


\bibitem[\protect\citeauthoryear{Yang and Kraut}{Yang and Kraut}{2017}]%
        {yang_persuading_2017-1}
\bibfield{author}{\bibinfo{person}{Diyi Yang} {and} \bibinfo{person}{Robert~E.
  Kraut}.} \bibinfo{year}{2017}\natexlab{}.
\newblock \showarticletitle{Persuading {{Teammates}} to {{Give}}:
  {{Systematic}} versus {{Heuristic Cues}} for {{Soliciting Loans}}}. In
  \bibinfo{booktitle}{\emph{Proceedings of {{CSCW}}}}.
\newblock


\bibitem[\protect\citeauthoryear{Yang, Wen, and Ros{\'e}}{Yang
  et~al\mbox{.}}{2015}]%
        {yang_weakly_2015}
\bibfield{author}{\bibinfo{person}{Diyi Yang}, \bibinfo{person}{Miaomiao Wen},
  {and} \bibinfo{person}{Carolyn~Penstein Ros{\'e}}.}
  \bibinfo{year}{2015}\natexlab{}.
\newblock \showarticletitle{Weakly {{Supervised Role Identification}} in
  {{Teamwork Interactions}}}. In \bibinfo{booktitle}{\emph{Proceedings of
  {{ACL}}}}.
\newblock


\bibitem[\protect\citeauthoryear{Zhang, Chang, {Danescu-Niculescu-Mizil},
  Dixon, Thain, Hua, and Taraborelli}{Zhang et~al\mbox{.}}{2018}]%
        {zhang_conversations_2018}
\bibfield{author}{\bibinfo{person}{Justine Zhang}, \bibinfo{person}{Jonathan~P.
  Chang}, \bibinfo{person}{Cristian {Danescu-Niculescu-Mizil}},
  \bibinfo{person}{Lucas Dixon}, \bibinfo{person}{Nithum Thain},
  \bibinfo{person}{Yiqing Hua}, {and} \bibinfo{person}{Dario Taraborelli}.}
  \bibinfo{year}{2018}\natexlab{}.
\newblock \showarticletitle{Conversations {{Gone Awry}}: {{Detecting Early
  Signs}} of {{Conversational Failure}}}. In
  \bibinfo{booktitle}{\emph{Proceedings of {{ACL}}}}.
\newblock


\bibitem[\protect\citeauthoryear{Zhang and {Danescu-Niculescu-Mizil}}{Zhang and
  {Danescu-Niculescu-Mizil}}{2020}]%
        {zhang_balancing_2020}
\bibfield{author}{\bibinfo{person}{Justine Zhang} {and}
  \bibinfo{person}{Cristian {Danescu-Niculescu-Mizil}}.}
  \bibinfo{year}{2020}\natexlab{}.
\newblock \showarticletitle{Balancing {{Objectives}} in {{Counseling
  Conversations}}: {{Advancing Forwards}} or {{Looking Backwards}}}. In
  \bibinfo{booktitle}{\emph{Proceedings of {{ACL}}}}.
\newblock


\bibitem[\protect\citeauthoryear{Zhang, Filbin, Morrison, Weiser, and
  {Danescu-Niculescu-Mizil}}{Zhang et~al\mbox{.}}{2019}]%
        {zhang_finding_2019}
\bibfield{author}{\bibinfo{person}{Justine Zhang}, \bibinfo{person}{Robert
  Filbin}, \bibinfo{person}{Christine Morrison}, \bibinfo{person}{Jaclyn
  Weiser}, {and} \bibinfo{person}{Cristian {Danescu-Niculescu-Mizil}}.}
  \bibinfo{year}{2019}\natexlab{}.
\newblock \showarticletitle{Finding {{Your Voice}}: {{The Linguistic
  Development}} of {{Mental Health Counselors}}}. In
  \bibinfo{booktitle}{\emph{Proceedings of {{ACL}}}}.
\newblock


\end{thebibliography}

\appendix
\end{document}